\documentclass[10pt,twocolumn,letterpaper]{article}

\usepackage{cvpr}              

\usepackage{graphicx}
\usepackage{amsmath}
\usepackage{amssymb}
\usepackage{booktabs}
\usepackage{tabularx}
\usepackage{color}
\usepackage{colortbl}
\usepackage{enumitem}
\usepackage{adjustbox}
\usepackage{wrapfig}

\definecolor{Gray}{gray}{0.9}
\definecolor{LightCyan}{rgb}{0.88,0.95,1}
\definecolor{blond}{rgb}{0.98, 0.94, 0.75}

\usepackage[pagebackref,breaklinks,colorlinks]{hyperref}

\newcommand\nnfootnote[1]{%
  \begin{NoHyper}
  \renewcommand\thefootnote{}\footnote{#1}%
  \addtocounter{footnote}{-1}%
  \end{NoHyper}
}

\begin{document}

\title{AI Art Neural Constellation: Revealing the Collective and Contrastive State of AI-Generated and Human Art}

\author{Faizan Farooq Khan*\\
KAUST \\
\and
Diana Kim*\\
KAUST\\
\and
Divyansh Jha \\
KAUST
\and
Youssef Mohamed\\
KAUST
\and
Hanna H Chang \\
KAUST
\and
Ahmed Elgammal \\
Rutgers University
\and
Luba Elliott \\
ELLUBA
\and
Mohamed Elhoseiny\\
KAUST
}

\maketitle

\vspace{-22pt}
\begin{abstract}
\vspace{-9pt}

 Discovering the creative potentials of a random signal to various artistic expressions in aesthetic and conceptual richness is a ground for the recent success of generative machine learning as a way of art creation. To understand the new artistic medium better, in this study, we conduct a comprehensive analysis to position AI-generated art within the context of human art heritage.  Our comparative analysis is based on an extensive dataset, dubbed ``ArtConstellation,''  consisting of annotations about art principles, likability, and emotions for 6,000 WikiArt and 3,200 AI-generated artworks. After training various state-of-the-art generative models, art samples are produced and compared with WikiArt data on the last hidden layer of a deep-CNN trained for style classification. We actively examined the various art principles to interpret the neural representations and used them to drive the comparative knowledge about human and AI-generated art—important artistic concepts in art history such as Wölfflin's and general art principles are used for interpretation. A key finding in the semantic analysis is that AI-generated artworks are visually related to the principle concepts for modern period art made in 1800 - 2000. In addition, through Out-Of-Distribution (OOD) and In-Distribution (ID) detection in CLIP space, we find that AI-generated artworks are ID to human art when they depict landscapes and geometric abstract figures, while detected as OOD when the machine art consists of deformed and twisted figures. We observe that machine-generated art is uniquely characterized by incomplete and reduced figuration. Lastly, we conducted a human survey about emotional experience. Color composition and familiar subjects are the key factors of likability and emotions in art appreciation. The diverse range of negative and positive emotions from AI-generated art to the comparable degree of human art is another observation from the survey. We propose our whole methodologies and collected dataset as our analytical framework to contrast human and AI-generated art, which we refer to as ``ArtNeuralConstellation''. The code and data are available at~\href{https://github.com/faixan-khan/ArtNeuralConstellation}{https://github.com/faixan-khan/ArtNeuralConstellation}
\end{abstract}
\nnfootnote{* indicates equal contribution.}\def\thefootnote{\arabic{footnote}}

\section{Introduction}
\label{sec:intro}

\begin{figure*}[t]
\centering
\includegraphics[width=1.0\linewidth]{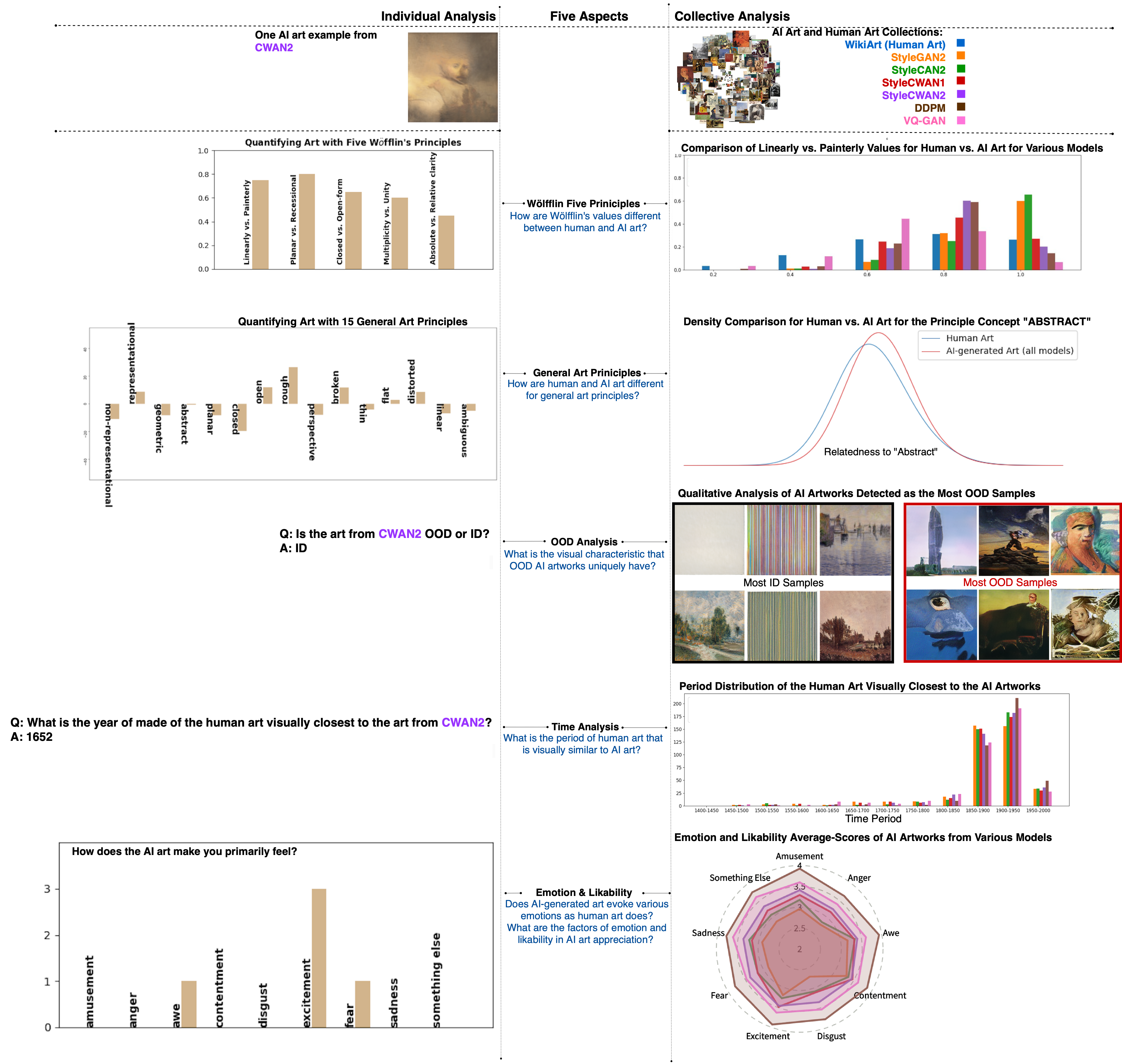}
\caption{This figure shows how our study reveals the collective state of AI-generated art within the context of human art heritage based on individual analysis from the five aspects: Wölfflin's concepts,  general art principles, Out-Of-Distribution analysis (OOD), time, and emotion \& likability.  The example bar plots and the answers in individual analysis present the actual results in our experiments for the AI portrait from the generative model CWAN2.}
\label{fig:main}
\end{figure*}

Recent great advancement in machine learning has affected the way of problem-solving by enabling cooperation with AI (Artificial Intelligence) in many domains; art is not an exception. The emergence of generative ML algorithms like GAN (Generative Adversarial Network)~\cite{gan_goodfellow} provides artists with a scientific tool to explore a new path for creativity more effectively than ever. Also, its artistic potential opens the various creative possibilities interpolating the two points: from pure randomness to authors' free will. 
\par Generative art is not a new genre; it existed as an art form since a long time ago in projects like Mozart's music generative system \textit{Musikalisches W\"{u}rfelspiel} (German for "musical dice game") in 1792. In terms of visual art, artists such as Harold Cohen have been developing computer systems with the ability to generate art autonomously since the 1970s.  New generative AI models are now seeing more attention from both the public and the art world along with a great amount of controversy, especially when machine learning is involved in art generation \cite{miller2019artist}. Given the nature of machine learning, some tricky issues can arise regarding authorship, copyright, computational creativity, etc. However, beyond all of that, we witness the unprecedented aesthetic and conceptual richness of machine learning arts and the opportunities in their high-dimensional hidden space to search for a new form of art never seen before; in $2018$, the first AI artwork (\textit{Portrait of Edmond de Belamy}) was sold at auction for over \$$400$K at Christie's New York. In  2022, an AI art generated by textual prompts (\textit{Th\'{e}\^{a}tre D'op\'{e}ra Spatial}) won first place in the Colorado State Fair Fine Art competition. In  2023, a prestigious museum MoMA in New York is exhibiting a large-scale installation by Refik Anadol using an AI model trained on modern art collections over the last 200 years; the name of the art is ``\textit{Unsupervised}".

At this fascinating moment for AI as a new, autonomous, and successful entity for art creation, it would be fundamental to understand how AI-generated art is different from the artworks artists have created. For comparative study, in this paper, we actively explore various art principles to show contrasting and comparable points between human and AI-generated art\footnote{In this paper, the terms ``human art" and ``AI-generated art" refer to the two end-poles indicating zero and 100\% of the intervention degrees of artificial automaton in art creation.}. Finally, we aim to promote a deeper understanding of AI technology as a new artistic medium, and furthermore, we hope the analytical protocol in this study provides guidance for the art and AI community to design and analyze the next-generation generative art. As a right step to basic, we focus on art principle concepts generally used in formal analysis of art history first.

Art styles, also referred to as art movements, have been central to studying art history, and they help differentiate between cultures, different periods, and inventions of new technologies; such inventions as paint tubes and photography, which were necessary for the development of the Impressionism and Cubism art movements, respectively, have heavily influenced art styles. The development of art movements also led to the development of different art principles, and theories have been developed to differentiate art movements. In this work, we apply the theory by Heinrich Wölfflin (1864-1945)~\cite{wolfflin1950principles}, which is considered one of the critical formulations of style patterns and style changes in art history. The comparative approach to formal analysis by Wölfflin has become a standard method of art history training~\cite{wolfflin_training}. To differentiate the Renaissance from the Baroque art style, Wölfflin used his method with five fundamental visual principles: (1) linearly vs. painterly; (2) planar vs. recessional; (3) closed-form vs. open-form; (4) multiplicity vs. unity; (5) absolute clarity vs. relative clarity.

Recent studies~\cite{elgammal2018shape, kim2018artprinciple, kim2022proxy} have assessed deep neural net representations from artwork using well-established artistic principles to understand how machines classify art styles. Elgammal et. al~\cite{elgammal2018shape} found that the last hidden space of a deep-CNN (Convolutional Neural Network) exhibits high correlations with Wölfflin's principles of art history, and~\cite{kim2018artprinciple, kim2022proxy} developed a deep neural network to learn a prototype space for style whose dimensions are aligned with 58 artistic concepts to describe the visual elements such as line, shape, color, texture, space, light, and subject. 
While W\"olfflin's classical principles are originally designed for discerning between Baroque and Renaissance styles~\cite{wolfflin1950principles}, the visual elements are the universal semantic structure which artists and art historians have used for effective communication and formal analysis in  art~\cite{taylor2014learning,ocvirk2002art}; we will call the elements ``general art principles" to differentiate them from Wölfflin's principles in this paper.

Inspired by the studies ~\cite{elgammal2018shape, kim2018artprinciple, kim2022proxy}, we conducted statistical analysis for the art principles after quantifying the art samples on the neural net's hidden layers to answer the following questions:  (1) how can we represent individual human and AI-generated art samples by the art principles?, (2) what are their collective states, so how does AI-generated art differ from human artwork?, and (3) can we detect generative art instances visually outlying from the general human art population? Through Out-Of-Distribution (OOD) detection, we specify distinctive visual features that AI-generated art uniquely has.  Pre-trained CLIP \cite{clip_radford} is used for the analysis as the base neural net representation, motivated by its strong capacity to encode fine-grained visual concepts in images.

Our contrastive evaluation is not limited to the formal analysis. Two other dimensions of time and emotions/likability are also investigated to identify a dominant period for human art samples that are visually closer to machine art. And, we also check if the machine generative art influences emotional experiences in art appreciation. Fig. \ref{fig:main} highlights the main philosophy of this study to focus on the collective state of machine art from the five dimensions after conducting individual analyses for massive human and machine art samples generated by various state-of-the-art models. Finally, we evaluate the difference between human and AI-generated art and there are the key findings as below.

\begin{itemize}
\item  \textbf{Wölfflin's art principles}: AI-generated artworks are related more to the concepts which are originally formulated to characterize the Baroque style by Wölfflin. 

\item \textbf{General art principles}: compared to human art, AI artworks are visually related more to the art principle concepts for modern art.

\item \textbf{Out-Of-Distribution (OOD) analysis}: human and AI-generated artworks are ID when they depict landscapes and geometric abstract figures, but when deformed and twisted figuration is generated by machines, the art samples are detected as OOD to human art. 

\item \textbf{Time analysis}:  The majority (70\%) of the AI-generated art was visually closest to the modern period art made from $1850$ to $2000$.

\item \textbf{Likability and emotional experience}: AI-generated artworks had higher likability when they were visually closer to or resembled human art. And, like human art, AI-generated artworks also evoke a diverse range of emotions. 
\end{itemize}

Along with the findings, we propose massive data collected for this study as a whole analytical framework to contrast human and AI-generated art. For 6,000 human art from the twelfth to twentieth century and 3,200 generated art pieces from eight different generative models, we collected annotations about Wölfflin's art principles, likability, and emotions resulting in 262,000 annotations for human and AI-generated art. The evaluated AI-generated artworks were generated using StyleGAN-based models with different creative loss functions—(1) the adversarial objective of the original StyleGAN \cite{karras2019style, Karras2019stylegan2}, (2) of Creative Adversarial Network (CAN) \cite{elgammal2017can}, and (3) of Creative Walk Adversarial Network (CWAN) \cite{stylecwan}, and two state-of-the-art generative models: Vector-Quantized GAN (VQ-GAN)~\cite{esser2021taming} and Denoising Diffusion Probabilistic Models (DDPM)~\cite{ho2020denoising}.

\section{Related Work}
\label{sec:related}
\subsection{Art Generation Models}
A main base model for art generation in our study is 
StyleGAN \cite{karras2019style,Karras2019stylegan2}. StyleGAN is an advanced model of Generative Adversarial Networks (GANs) \cite{gan_goodfellow} that achieves high-quality image generation through the architecture feeding stochastic variations to different layers which results in hierarchical controls from global style to fine details in local visualization. Theoretically, GANs~\cite{gan_goodfellow,radford2015unsupervised} can learn visual distributions in training data and use them to synthesize outputs similar to the training data.

Art creation is the domain that substantially benefits from creative generation. The literature on art creation demonstrates how GAN algorithms have shifted from imitative to creative by modifying the GAN loss encouraging the production of a novel, but also semantically meaningful, content ~\cite{elgammal2017can, stylecwan, paintingGAN, hertzmann_creative}. In this study, we employ such networks to allow machines to be creative by training StyleGAN architectures with the creative objectives in \cite{elgammal2017can,stylecwan}.

Along with the GAN models, two state-of-the-art image synthesis models are examined in our study. In recent advancements in image synthesis, notable approaches such as VQ-GAN~\cite{esser2021taming} and DDPM~\cite{ho2020denoising} have emerged as powerful alternatives to traditional GANs. VQ-GAN leverages the transformative capabilities of transformers~\cite{vaswani2017attention}, initially designed for language tasks, to produce high-resolution images characterized by intricate details. DDPM presents a generative model that progressively converts noise distributions into target data distributions, enabling the synthesis of high-quality images.

\subsection{Formal Analysis of Art in Neural Nets}
In our study, art samples are encoded in a deep neural net, and the quantitative representations are interpreted with art principles to derive semantic knowledge about AI-generated art. To understand the right positioning of the generative models within the context of human art heritage, the deep neural nets are trained by classification for 21 human art movements. Our approaches are inspired by the previous works of \cite{elgammal2018shape, kim2018artprinciple} which showed the ability of deep-CNNs to learn a smooth visual transition over time in their last hidden layers when the neural nets are trained for style classification. By interpreting the hidden space with Wölfflin's and general art principles \cite{wolfflin1950principles, kim2022proxy}, they demonstrated some art historical facts and theories and proved the potential of deep neural nets as a computational framework for formal analysis of visual art \cite{elgammal2018shape, kim2018artprinciple}. To the best of our knowledge, our work is the first exploration using art principles to compare human and AI-generated pieces together in a neural net space.

Our neural net analysis includes general semantics beyond the art principles by considering CLIP space~\cite{clip_radford}. We observed a complete separation between human and AI-generated art in CLIP vision space, so we conducted an OOD analysis. OOD is an important research subject for safe and reliable machine learning \cite{fort2021exploring, tao2023non, sun2022knnood, du2022vos}. We need to hold a final AI decision when inputs are statistically far from the original training data to avoid abnormal operations. Detecting the outlying samples well is the principal goal of OOD methods, but in this work, we use it differently from the original purpose; we deployed the nearest neighbor OOD detection method \cite{sun2022knnood} to detect AI-generated samples whose nearest neighbor samples of WikiArt are notably distant. 

\subsection{Emotional Analysis}
Emotional response is a central aspect in art appreciation and aesthetics \cite{silvia2005emotional}; it has been an intriguing subject for the deep learning and computational art community to find the relation between visual features and emotion \cite{cetinic2019deep} and to develop an automatic system to predict emotions for input paintings \cite{mohammad2018wikiart, campos2017pixels, achlioptas2021artemis}. Most recently, Achlioptas et al. \cite{achlioptas2021artemis} introduced ArtEmis, a large-scale dataset of emotional reactions to visual artwork joined with explanations of these emotions in language, and developed machine learning models for dominant emotion prediction from images or text. In our study, we conduct a human survey to observe how people construct emotions from AI-generated art in contrast to human art. The newly collected annotations 
for AI-generated art extend ArtEmis. The labels and explanations about subjects' emotions follow the same emotional categories and survey questions in ArtEmis.

\section{Experimental Setup}
\label{sec:experiments}

The various state-of-the-art models of neural nets and semantic spaces are tested inclusively to set our experiment bias-free. For image generation, several variants of StyleGANs  \cite{karras2019style,Karras2019stylegan2} are trained and ready along with VQ-GAN \cite{esser2021taming} and DDPM  \cite{ho2020denoising}. All the models are trained from scratch on the WikiArt dataset. For art comparison, the four domains of (1) Wölfflin's and (2) general art principles, (3) time, and (4) emotions are considered. For neural nets,  all representative CNNs: VGG-Nets and ResNets, and vision transformers are trained as style classification for art principle analysis, and a CLIP (ViT-B/32) model is also adopted as the space for deep nearest neighbor OOD analysis \cite{sun2022knnood}. 

\subsection{Artworks by Human Artists and Machines}

\subsubsection{Models for Machine Art Generation}
In this paper, we build six GAN-type models for art generation—based on the two versions of StyleGAN-1 and StyleGAN-2~\cite{karras2019style, Karras2019stylegan2}, three objectives are combined (1) the adversarial objective of the original StyleGAN, (2)  Creative Adversarial Network (CAN) \cite{elgammal2017can}, and (3) Creative Walk Adversarial Network (CWAN) \cite{stylecwan}. There exist various GAN models, but the choice of StyleGAN is for its superior performance in image generation. Our models are not limited to StyleGANs. We include a diffusion model,  DDPM~\cite{ho2020denoising}. DDPM-based models use diffusion processes to progressively transform a noise distribution into a target data distribution allowing for high-quality image synthesis. Our last model is VQ-GAN \cite{esser2021taming} which combines a Variational Autoencoder (VAE)~\cite{kingma2022autoencoding} with a Vector Quantization layer and uses a transformer to produce high-quality and diverse image samples. In summary, we trained eight models for machine art generation: StyleGAN-1\&2, StyleCAN-1\&2, StyleCWAN-1\&2, DDPM, and VQ-GAN. In the later sections,  will bold the original Style\textbf{GAN}-1\&2  to avoid confusion with StyleCAN-1\&2. All models are trained by using the WikiArt dataset (https://www.wikiart.org). 

\subsubsection{Data Setup}\label{group}
Two sets of art data are prepared.\\
\noindent \textbf{Machine-Generated Art:} we generated 10,000 images from the eight trained Style\textbf{GAN}, StyleCAN, StyleCWAN, DDPM, and VQ-GAN models and sampled four groups from each of the models; each group contains 100 generated pieces of art, so in total, we have 3,200 machine-generated artworks.
\begin{enumerate}
    \item Highest Nearest Neighbour (NN): we computed the NN distance between AI-generated and human art from the WikiArt dataset. Then, we selected the top 100 machine art with the highest NN distance.
    \item Lowest NN: we selected another 100 pieces of art with the lowest NN distance. 
    \item Highest shape entropy: we selected the artworks with the highest confusion in art style computed from a trained style classifier.
    \item Random: we selected a set of 100 random images.
\end{enumerate}

\noindent  \textbf{Digital Human Art Heritage:}
for art data drawn by human artists, we selected  1,000 samples from each century from 1100 to 2000, 6,000 in total. All these images are Western art, except for Ukiyo-e in 17th-19th, and they are collected from WikiArt. 

\subsection{Semantic Spaces for Contrastive Analysis}
\subsubsection{Wölfflin's Principles} \label{data_wofflin}
We conducted our analysis for the five Wölfflin's principles \cite{wolfflin1950principles}. Even though Wölfflin proposed them initially to differentiate Renaissance and Baroque styles, we posit that they would be applicable to our task to capture the visual difference between human and AI-generated art. The principles consist of the five pairs of opposed or contrary concepts as below.
\begin{enumerate}
    \item \textbf{linearly vs. painterly:} paintings with linearly characteristics depict isolated objects and clear boundaries where all the figures are illuminated. Painterly paintings depict blurry outlines and swift brushstrokes.
    \item \textbf{planar vs. recessional:} paintings with planar characteristics have objects arranged in planes parallel to the plane of the canvas. In recessional paintings, these objects can be in angle and focus on spatial depth.
    \item \textbf{closed-form vs. open-form:} all figures in closed-form are balanced within the frame, and the main subject is easily distinguishable. In contrast, the figures are cut off in open-form, and there is no single focus. Although the former is primarily self-contained, the latter indicates space beyond the frame.
    \item \textbf{multiplicity vs. unity:} paintings with multiplicity characteristics have distinguished parts, and each part demonstrates independent features. While in unity, elements are not viewed as individuals but a coherent part of the entire painting. The figures weld together, and colors blend in.
    \item \textbf{absolute clarity and relative clarity:} absolute clarity has a realistic representation enabling the viewer to understand the form of the objects. In contrast, relative clarity enhances representations with visual effects, viewed as a whole.
\end{enumerate}

We collected data related to Wölfflin's principles by training people to learn about a specific concept of Wölfflin's principles and identify a painting based on the learned concepts. The survey interface provided descriptions of Wölfflin's principles (one at a time) to the annotators. Based on the explanation of linearly and painterly paintings, the viewer selected a rating scale from 1 to 5 (1: clear Linearly, 2: mostly linearly,  3: borderline, 4: mostly painterly, 5: clear painterly). We collected five individual responses per painting for each of Wölfflin's principles. These five responses further help accurately assess a particular principle. Then, we averaged the five ratings for every piece of artwork and normalized the resulting score between 0 and 1, where normalized scores closer to 0 are considered linear, and scores closer to 1 indicate painterly. The interface is attached to Appendix~\ref{data_c_supp}.

\subsubsection{General Art Principles} \label{general art principles}
Along with Wölfflin's five principles,  general art principles such as line, shape, color, texture, and space  are also considered. A feasible framework for the analysis would be proxy-space~\cite{kim2018artprinciple,kim2022proxy}. Proxy space is a deep neural network trained to quantify the relatedness of input paintings to 58 art principle concepts in its last hidden layer, while the whole network is trained for style classification by using style labels, paintings, and the general relationship between 58 visual concepts and 20 styles as training data. In proxy space, each dimension is readily aligned with a semantic, so statistical analysis for human and generated artworks is directly available.

We adopted the same model from the original framework~\cite{kim2022proxy} with an optional choice—estimation of the general relationship between visual concepts and styles by the sample mean of ground truth values—and selected a subset among 58 visual concepts based on the reported \textbf{A}rea \textbf{U}nder the receiver operating characteristic \textbf{C}urve (AUC) performance (more than 0.75 AUC). The concepts used in our statistical analysis are listed as follows: (1) non-representational, (2) representational, (3) geometric, (4) abstract, (5) planar, (6) closed, (7) open, (8) rough, (9) perspective, (10) broken, (11) thin, (12) flat, (13) distorted, (14) linear, and (15) ambiguous.

\subsubsection{Out-Of-Distribution (OOD) Analysis}
GAN models are theoretically designed to learn real art distribution, but in practice, due to the limitations in architectures, learning algorithms, and data, AI-generated art looks quite different from human art even to our naked eyes sometimes.

Can a machine space help us to identify outlier samples and direct us to see how the machine and human art are visually similar or dissimilar? A qualitative and comparative examination is performed based on the results from OOD analysis; we detect OOD and ID instances of the machine generative art compared to human art and qualitatively analyze what features determine ID or OOD. CLIP \cite{clip_radford} vision space is used in the OOD analysis for its strong capacity to encode profound textural information.

\subsubsection{Time Analysis}\label{time}
Deep-CNN style classifiers can encode the smooth visual transition of art over time in its hidden layers. Even though time information is not used in training, the hidden representation learned a positive correlation with the year of art creation \cite{elgammal2018shape}. Based on the previous finding, we estimate the time period of AI-generated art by checking the year of the human art which is spatially closest to each of AI artworks in a  neural space.

\subsubsection{Emotion and Likability}\label{emotion and likability}
Likability and emotion are important aspects to reflect the beholder's subjective visual experience of art. In this paper, we set up an experiment to answer the following specific questions: is AI-generative art more liked or disliked?  What visual features can we notice from the art people like the most?  Will generative artworks bring diverse emotions as human art does? If it does, how can we interpret the results?

For emotional annotation of human art, we make use of ArtEmis data ~\cite{achlioptas2021artemis, youssef2022artemis2}; WikiArt paintings are labeled by one of the nine emotions: (1) amusement, (2) awe, (3)  contentment, (4) excitement, (5) anger, (6) disgust, (7) fear, (8)  sadness, (9)  something else.
To collect emotional data for AI-generative art, we asked our Mechanical Turk participants two questions on the survey: Q1 ``How does this painting make you primarily feel?'' We displayed the artwork to the participant, who selected an emotion. After choosing the emotion, the participants were asked to explain why they felt that particular emotion: Q2 ``Give a detailed description (at least eight words) about why the participant feels like this, based on specific details of the painting.'' We collected this data for all $3,200$ AI-generated artworks, and five responses per artwork; it resulted in $16,000$ responses.

For likability analysis, we followed the same likability and Turing experiments in \cite{elgammal2017can}. We asked the participants two questions. Q1 ``How much do you like this image?'' The user selected a rating from 1 to 5 (1: very much dislike to 5: very much like). We also asked Q2 ``Whether the art is created by an artist or a machine?'' (Turing Test). This data is also collected for all 3,200 AI-generated artworks.

\subsection{Deep Neural Nets}
\cite{elgammal2018shape} is a seminal work for computational formal analysis for visual art. It showed that the hidden space of a deep-CNN style classifier is capable of representing the visual transitions by art movements; (1) real values on the space are correlated with certain visual concepts for art (Wölfflin's principles), and (2) arts' smooth visual transitions over time can be explained by the concepts. Inspired by the finding, in this paper, we trained various deep neural nets to classify style and analyzed Wölfflin's principles and time from the hidden space of the neural nets. 

The last hidden space after ReLU activation functions (Rectifier Linear Unit) is used to collect our neural net representations. Then, we applied PCA (Principal Component Analysis) to the features and examined the principal components, which cover 95\% of the variance. To select a model for our analysis, style accuracy and Pearson's correlation coefficients (PCC) with time are assessed as a preliminary, following the experiments in \cite{elgammal2018shape}. We tested various convolutional and transformer-based networks such as VGG-Net16, ResNet50, ResNet10,  and ViT-S, B, and L. Pretrained and fine-tuned ResNet50 architectures achieved the highest art-style classification accuracy and showed the maximum correlation with time. Therefore, we choose ResNet50 for further discussion in the results. The detailed results of style accuracy and time correlation for different models are reported in Appendix~\ref{class_supp}.

\section{Experimental Results}
\label{sec:results}
\begin{figure}[t]
    \centering
    \includegraphics[width=1.0\linewidth]{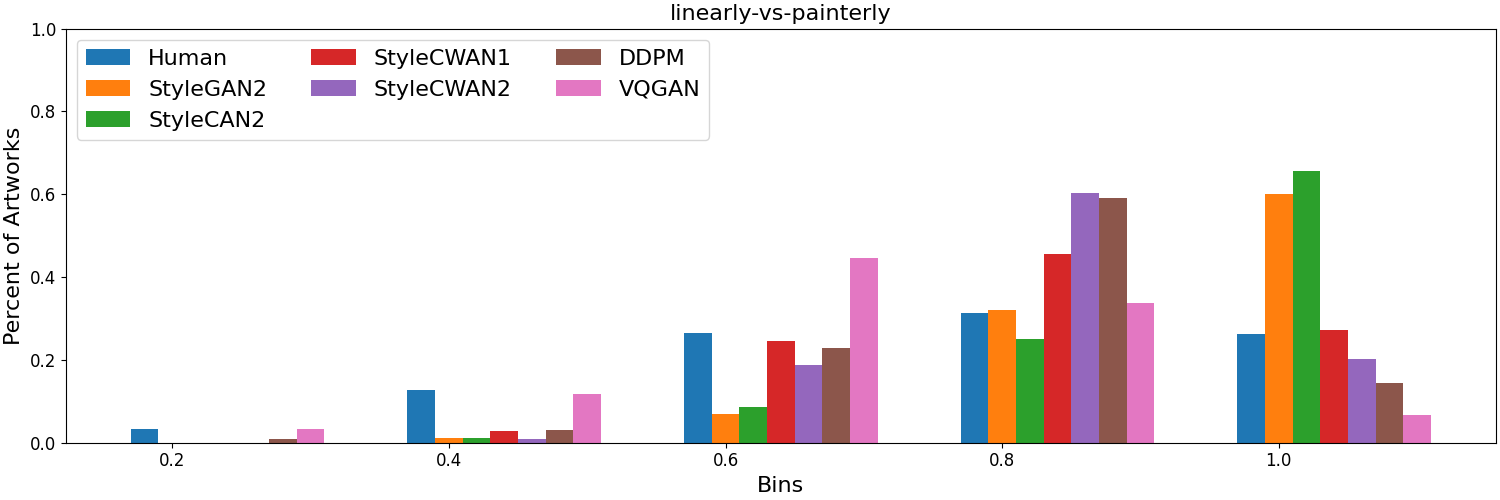}
    \caption{The distribution example of the first principle: linearly vs. painterly. The bins represent the values from the left value to the right value.  As we can see in this figure, AI-generated art is less uniform than human art for  Wölfflin's principles. Machine artworks are highly populated on the right side concepts more in general. More distribution figures for other principles are shown in Appendix~\ref{expaned_woff_supp}. }
    \label{fig:ratio_wofflin}
\end{figure}

\subsection{Wölfflin's Principles-based Analysis}
We reconfirmed the finding in \cite{elgammal2018shape} that Wölfflin’s principles are implicitly learned in the last hidden layer of CNN style classifiers. As an example result, the PCCs between Wölfflin's concepts and ResNet50 features for WikiArt art are (1) linearly-painterly: $-0.21$, (2) planar-recessional: $-0.19$, (3) closed-open: $0.2$, (4) multiplicity-unity: $0.25$, (5) absolute-relative clarity: $0.33$. The results are the maximum absolute PCC among the top 30 principal components (95\%). Almost no correlation is found after the 30th component. 

\begin{table}[t]
\begin{center}
\caption{PCC of various architectures on human and AI-generated art: The table presents the averaged values of absolute maximum PCCs of the first 30 PCA components—for all five principles. The highest PCC for each type of art is highlighted in bold, followed by the second highest in red color. Human art consistently has higher PCC than AI art for all of Wölfflin's principles across all the architectures. The models with ``+2" are extended  ConvNet models adding the two hidden layers of 1024 and 512-dimensional nodes on the top of the original models.}
\label{tb:coefficient}
\scalebox{0.65}{
\begin{tabular}{l|c|c|c|c|c|c|c}
			\toprule
			Architecture & Human Art & SG2 & SC2 & CW1 & CW2 & DDPM & VQ-GAN\\
			\midrule
			ResNet50   &	\textbf{.236} &	.114 &	.126 &	\textcolor{red}{.128} 	&	.124 &	.126 &	.122	\\ 
			ResNet50+2  &	\textbf{.186} &	.114 &	.084 &	\textcolor{red}{.130} 	&	.118 &	.096 &	.122	\\ 
			ResNet101   &	\textbf{.262} &	.122 &	.106 &	.108	&	\textcolor{red}{.128} &	.100 &	.116	\\ 
			ResNet101+2 &	\textbf{.170} &	.096 &	.094 &	\textcolor{red}{.134}	&	.116 &	.092 &	.118	\\ 
            VGG16&	\textbf{.300} &	\textcolor{red}{.144} &	.110 &	.112	&	.134 &	.108 &	.126	\\ 
			VGG16+2&	\textbf{.344} &	.118 &	.112 &	.116	&	\textcolor{red}{.124} &	.108 &	.122	\\ 
			ViT-S&	\textbf{.242} &	.110 &	.108 &	.142	&	.102 &	.118 &	\textcolor{red}{.146}	\\ 
			ViT-B&	\textbf{.256} &	\textcolor{red}{.136} &	.122 &	.120	&	.110 &	.102 &	.116	\\ 
			ViT-L&	\textbf{.224} &	\textcolor{red}{.142} &	.110 &	.124	&	.132 &	.116 &	.136	\\ 
			\bottomrule
\end{tabular}}
   
\end{center}
\end{table}

Human and AI-generated art are compared by Wölfflin’s principles. For all neural nets and art models, in Table~\ref{tb:coefficient}, we averaged the maximum absolute PCC in the top 30 principal components for the five Wölfflin's principles; we observed that the correlation for human artworks is consistently higher than AI-generated images. Given the fact that the results are based on absolute correlation, the less-valued PCC indicates  AI-generated art is varied within the smaller ranges of the two conceptual poles in   Wölfflin's principles than human art.

We confirm the smaller variation of AI art again in additional experiments. First, in Table \ref{tb:ratio_wofflin}, for all principle concepts, AI-generated art's standard deviations are less than human art's. In addition, as shown in Fig. \ref{fig:ratio_wofflin}, machine art is more distant from uniform than human art, but instead highly populated on the high values around $0.4$ in general. Relatively, the concepts in the left poles, originally selected for Renaissance-style characterization, are sparsely observed in machine generative art. Instead, the mean values of AI-generated art are more biased toward the four visual concepts in the right poles: ``painterly, relative clarity, unity, open" while the deviation is smaller than human art across all the principles, forming certain visual characteristics. This phenomenon may be related to the results in the later sections; the AI-generated art is visually close to modern period human art drawn in 1800 - 2000.

\begin{table}[t]
\begin{center}
\caption{Means and standard deviations for each of Wölfflin's principles for human and AI-generated art.}
\label{tb:ratio_wofflin}
    
\scalebox{0.6}{
    \begin{tabular}{@{}l|c|c|c|c|c@{}}
    \toprule
        & \textbf{Linearly}& \textbf{Planar} & \textbf{Closed-Form} & \textbf{Multiplicity} & \textbf{Absolute Clarity} \\
        & \textbf{vs}& \textbf{vs} & \textbf{vs} & \textbf{vs} & \textbf{vs} \\ 
        & \textbf{Painterly }& \textbf{Recessional} & \textbf{Open-Form} & \textbf{Unity} & \textbf{Relative Clarity} \\ \midrule
        Human Art & 0.61/0.22 & 0.52/0.19 & 0.60/0.21 & 0.56/0.21 & 0.58/0.28 \\ 
        Style\textbf{GAN2} & 0.79/0.14 & 0.53/0.17 & 0.63/0.13 & 0.62/0.15 & 0.71/0.17 \\ 
        StyleCAN2 & 0.80/0.15 & 0.54/0.14 & 0.61/0.13 & 0.65/0.11 & 0.70/0.14 \\ 
        StyleCWAN1 & 0.66/0.14 & 0.66/0.13 & 0.59/0.15 & 0.59/0.14 & 0.62/0.13 \\ 
        StyleCWAN2 & 0.68/0.12 & 0.64/0.12 & 0.64/0.12 & 0.63/0.13 & 0.69/0.13 \\ 
        \bottomrule
    \end{tabular}}
    
\end{center}
\end{table}

\begin{figure*}[t]
\centering
\includegraphics[width=1.0\linewidth]{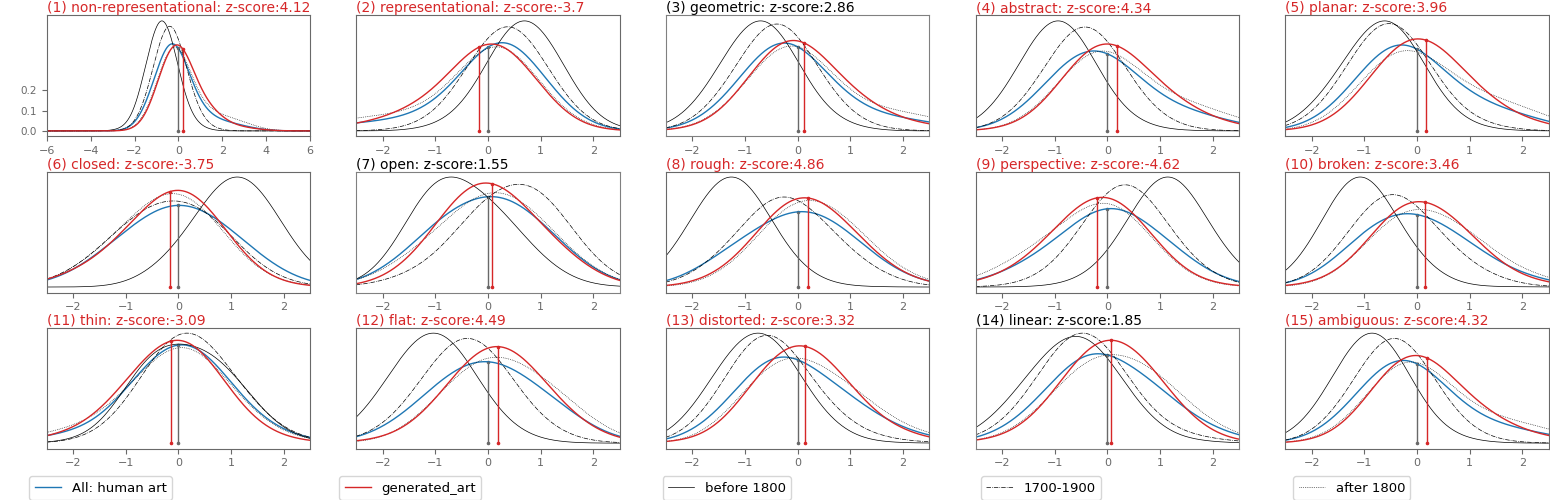}
\caption{For each visual concept, the densities for human (blue lines) and generated art (red lines) are estimated to contrast their centers. Dotted densities are also added to show how human artworks visually changed from classical to modern times and to see where the generated art falls on the historical transitions. The plots' ranges are magnified around centers, and the names of the concepts are colored red when the two population means are significantly different. We found AI-generated art centers are different from the human centers for most of the visual concepts except for ``geometric", ``open", and ``linear";  in general, generated artworks are visually biased toward modern periods artworks. }
\label{fig:density}
\end{figure*}

\begin{table*}[ht!]
\begin{center}
\caption{After dividing human art samples into two groups by the year of made: before and after $1800$, we computed $Z$-statistics to see how  AI paintings are different from the time groups. We observed that AI art is visually similar to modern art while they are significantly different from classical human art.}

\begin{adjustbox}{width=\textwidth}
\begin{tabular}{c| c|c|c|c|c|c|c|c|c|c|c|c|c|c|c}
\toprule
& \multicolumn{15}{c}{\textbf{ $Z$- statistics for 15 Visual Concepts}} \\ \midrule
\textbf{years} & non-representational & representational & geometric & abstract & planar 
& closed & open & rough  & perspective & broken 
& thin & flat & distorted & linear & ambiguous \\ \midrule
        
$-1800$ & 19.90 & -18.39 &19.07 & 25.99 & 19.60 & -28.70 & 12.34 & 34.75& -31.49 & 28.35 & -6.52 & 29.58 & 21.50 & 17.96 & 24.01\\ 
$1800-$ & -1.21  & 1.39 & -2.66 & -2.27 &-1.37 & 1.24 & 0.22 & -2.24 & 2.78 &-3.02 & -0.79 & -2.38 & -1.86& -3.0 & -1.94\\ 
\bottomrule
\end{tabular}
\end{adjustbox}
\label{tb:before_after_1800}
\end{center}
\end{table*}

\subsection{General Art Principles based Analysis}\label{general art principles_experiment}
To compare human and AI-generated artworks by general art principles, we reproduced the same neural network from the original framework~\cite{kim2022proxy} to quantify all paintings with their relatedness to each of the 15 visual concepts. For the unified scale, we normalized the AI and human embeddings (\#samples $\times$ 15) by the 15 means and 15 standard deviations of human artworks and compared how the AI and human artworks are statistically different for each concept—we will call the normalized embeddings ``standardized values by human art". On the normalized space, Kernel Density Estimation (KDE) and hypothesis testing are conducted to decide whether AI and human art samples have the same population means or not. The detailed experimental procedure is reported in Appendix~\ref{proxy-analysis_supp}.

The densities in Fig.~\ref{fig:density} provide overviews of how human and generated artworks are different by which visual concept and how much the difference is significant. To plot Fig.~\ref{fig:density}, all eight models' generated samples are combined to get a single representative density of generated art for each concept,  and a Gaussian kernel with $\sigma  = 0.5$ is used for KDE. In hypothesis testing, we found that the generated centers are different from the human centers for most of the visual concepts except for ``geometric", ``open", and ``linear" ($p$-value: $0.0026$). The test statistic is $Z=\frac{\bar{X_{g}} -\bar{X_{r}}}{\frac{S_{g}}{\sqrt{n}}}$, where $\bar{X_{g}}$ and $\bar{X_{r}}$ are respectively the sample means for generated and human artworks, and $S_{g}$ is the sample standard deviation of generated samples, and $n$ is the number of samples.

Another observation to remark on is that the generated arts are visually biased toward modern periods' human art. When we computed KDE after dividing the human samples into three groups by the periods: (1) before $1800$, (2) $1700 - 1800$, and (3) after $1800$, we observed definite linear transitions of human art for visual concepts across time (dotted and black lines in Fig.~\ref{fig:density}). And, the visual bias between generated and human art is also aligned with the direction of the changes. For example, as human art is getting less ``representational" as closer to the modern period, the generated art is less ``representational" compared to human art; the same phenomenon is observed for all the visual concepts. To confirm this observation, we tried the same hypothesis test by dividing human samples into two groups: before $1800$ and (3) after $1800$ in Table~\ref{tb:before_after_1800}; the center of generated samples is significantly different from the human arts as human art samples are drawn before the 1800s while the center of modern art is not much different from generated art's. The results show that generated art is visually biased to the modern period art from the perspectives of the principle concepts; it is consistent with the result in Fig.~\ref{resnet-bar}. The tendency is observed again for individual  AI models. For each model, we observed the visual resemblance between AI and human modern art again even though the models had different significant concepts. Testing results for individual models are in Appendix~\ref{hypo_test_supp}.

To check actual instances on the proxy space, we sorted generated samples based on the negative logarithm of the tail probability on the normalized space by human art—as further from the center of human arts, generated arts will be valued more on the negative logarithm. We will name the metric ``OODness-proxy (Out-Of-Distribution)". Through the instance observation from the OODness orders,  we aimed to answer the questions: (1) what are the nearest and farthest generated samples to human art, (2) how do they look depending on the distances, and (3) are the farthest samples also on the same side of the bias to modern art as we observed in our hypothesis testing? (4) Lastly, we check whether the generated arts are valued within the range of human arts or beyond on each axis of 15 visual concepts.

\begin{table*}
  
\centering
   \begin{adjustbox}{width=0.95\textwidth}
   \begin{tabular}
   {c|c|cc|cc|c}
   \toprule
   &human&\multicolumn{4}{c|}{AI-generated}&human\\\midrule
   & most in-lied human&\multicolumn{2}{c|}{most in-lied AI-generated}&\multicolumn{2}{c|}{most out-lied AI-generated}& most out-lied human\\\midrule
Non-representational&\includegraphics[height=0.135\textwidth, width=0.135\textwidth]{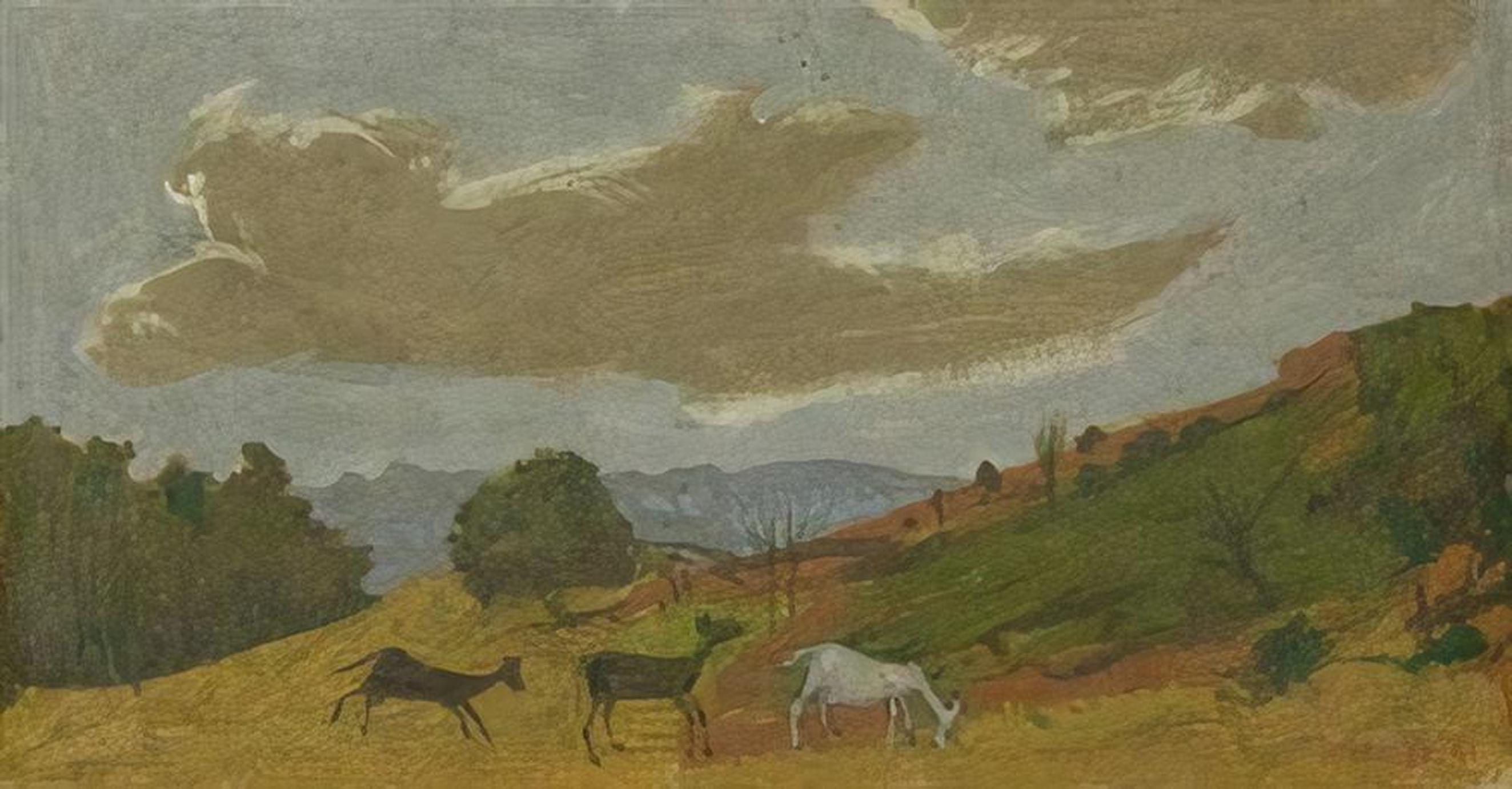}
   &\includegraphics[height=0.135\textwidth, width=0.135\textwidth]{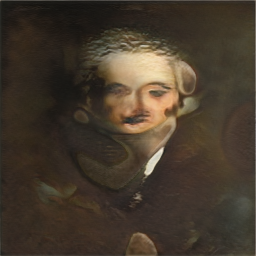}
   &\includegraphics[height=0.135\textwidth, width=0.135\textwidth]{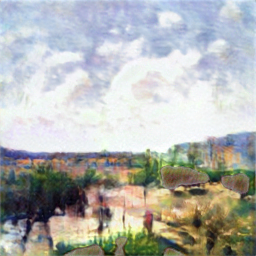}
   &\includegraphics[height=0.135\textwidth, width=0.135\textwidth]{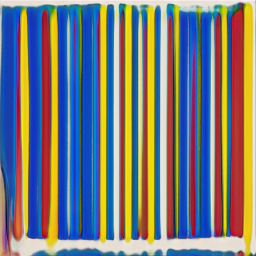}
    &\includegraphics[height=0.135\textwidth, width=0.135\textwidth]{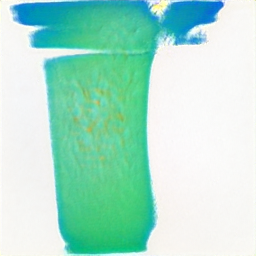}
     &\includegraphics[height=0.135\textwidth, width=0.135\textwidth]{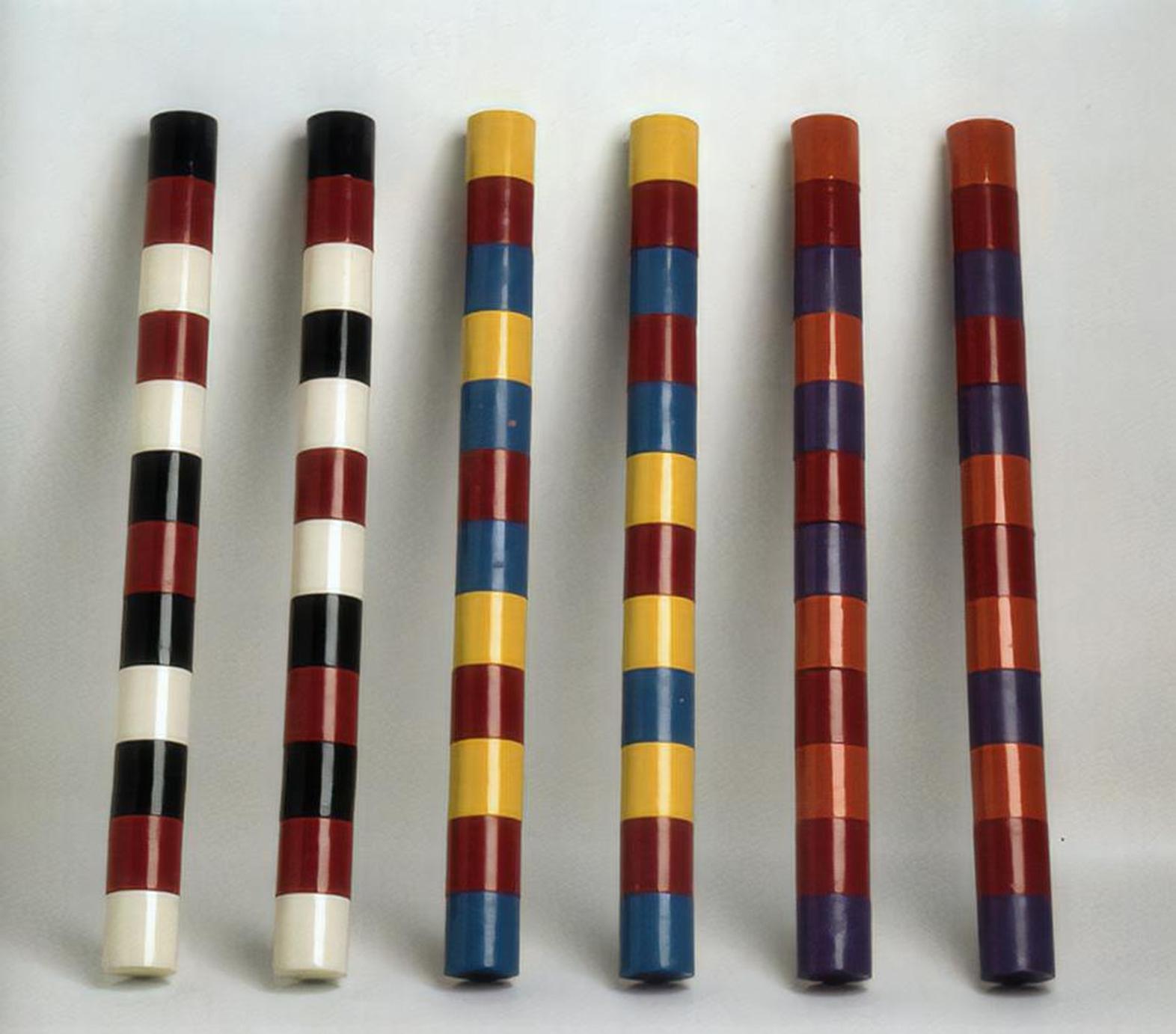}\\
     (scores)&($0.69,0.0$)&($0.69,0.0$)&($0.69,0.0$)&($13.09,4.61)$&($13.56,4.70$)&($24.97,6.65$)\\\midrule

abstract&\includegraphics[height=0.135\textwidth, width=0.135\textwidth]{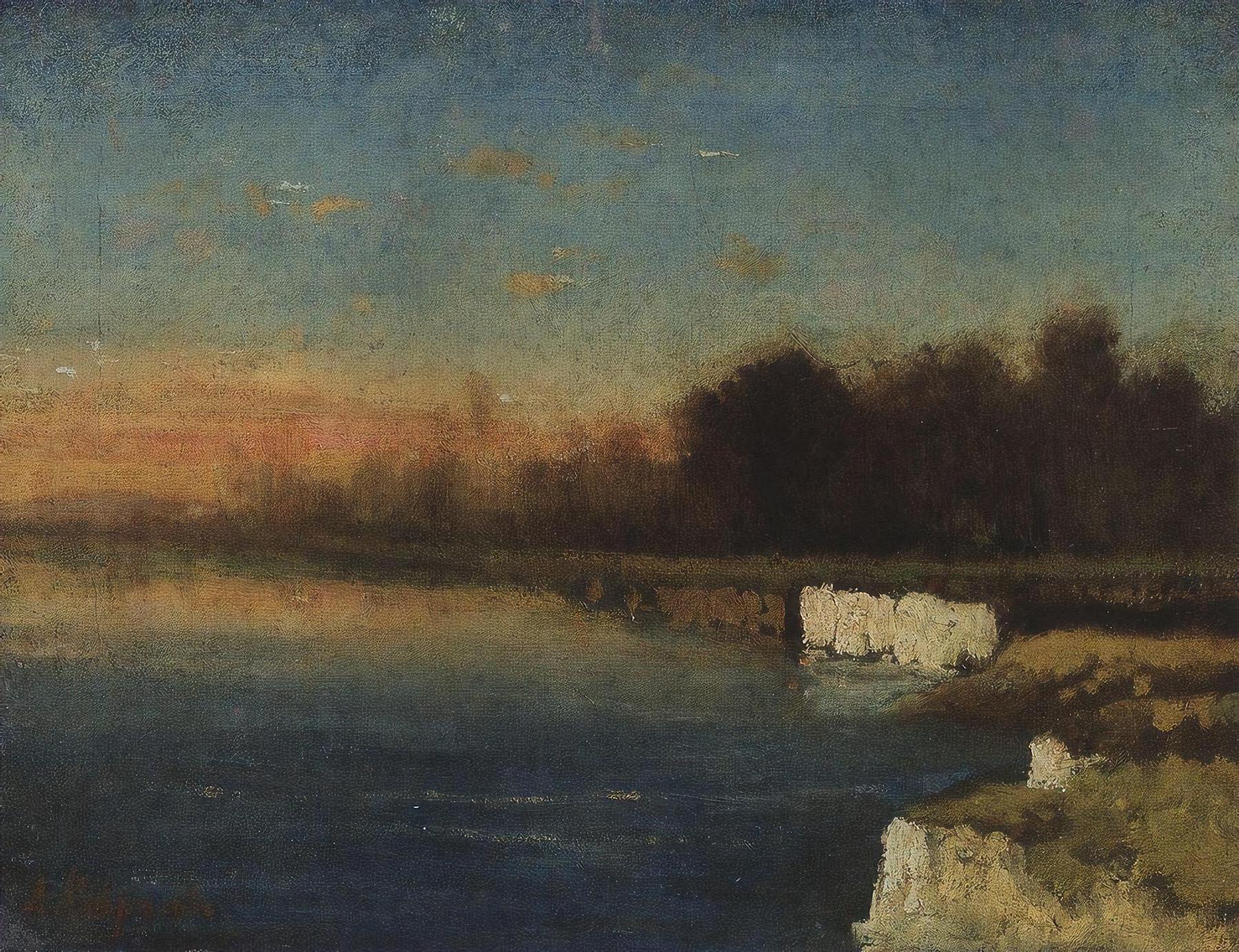}
&\includegraphics[height=0.135\textwidth, width=0.135\textwidth]{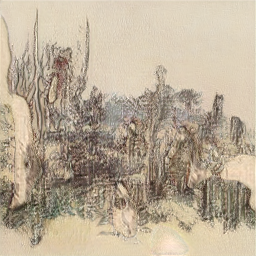}
&\includegraphics[height=0.135\textwidth, width=0.135\textwidth]{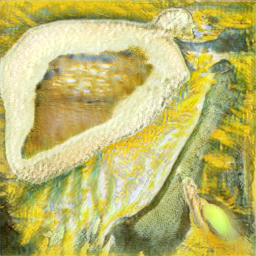}
&\includegraphics[height=0.135\textwidth, width=0.135\textwidth]{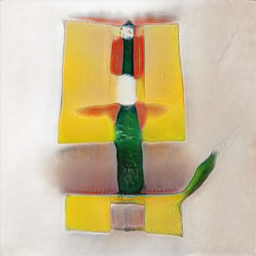}
&\includegraphics[height=0.135\textwidth, width=0.135\textwidth]{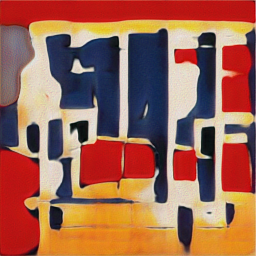}
&\includegraphics[height=0.135\textwidth, width=0.135\textwidth]{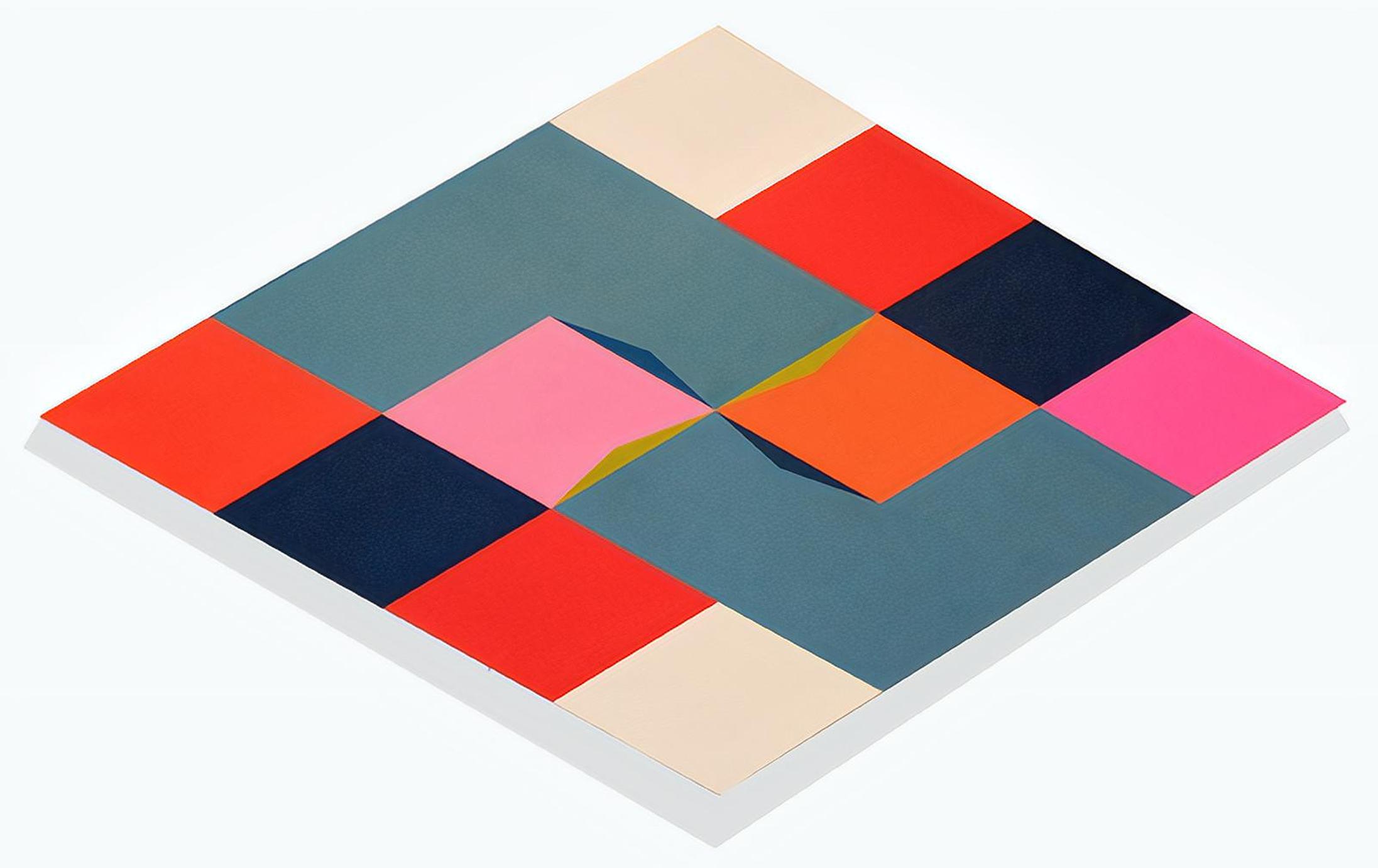}\\
(scores)&($0.69,0.001$)&($0.69,0.0$)&($0.69,0.0$)&($9.83,3.87$)&($11.20,4.19$)&($15.59,5.10$)\\\midrule

     closed&\includegraphics[height=0.135\textwidth, width=0.135\textwidth]{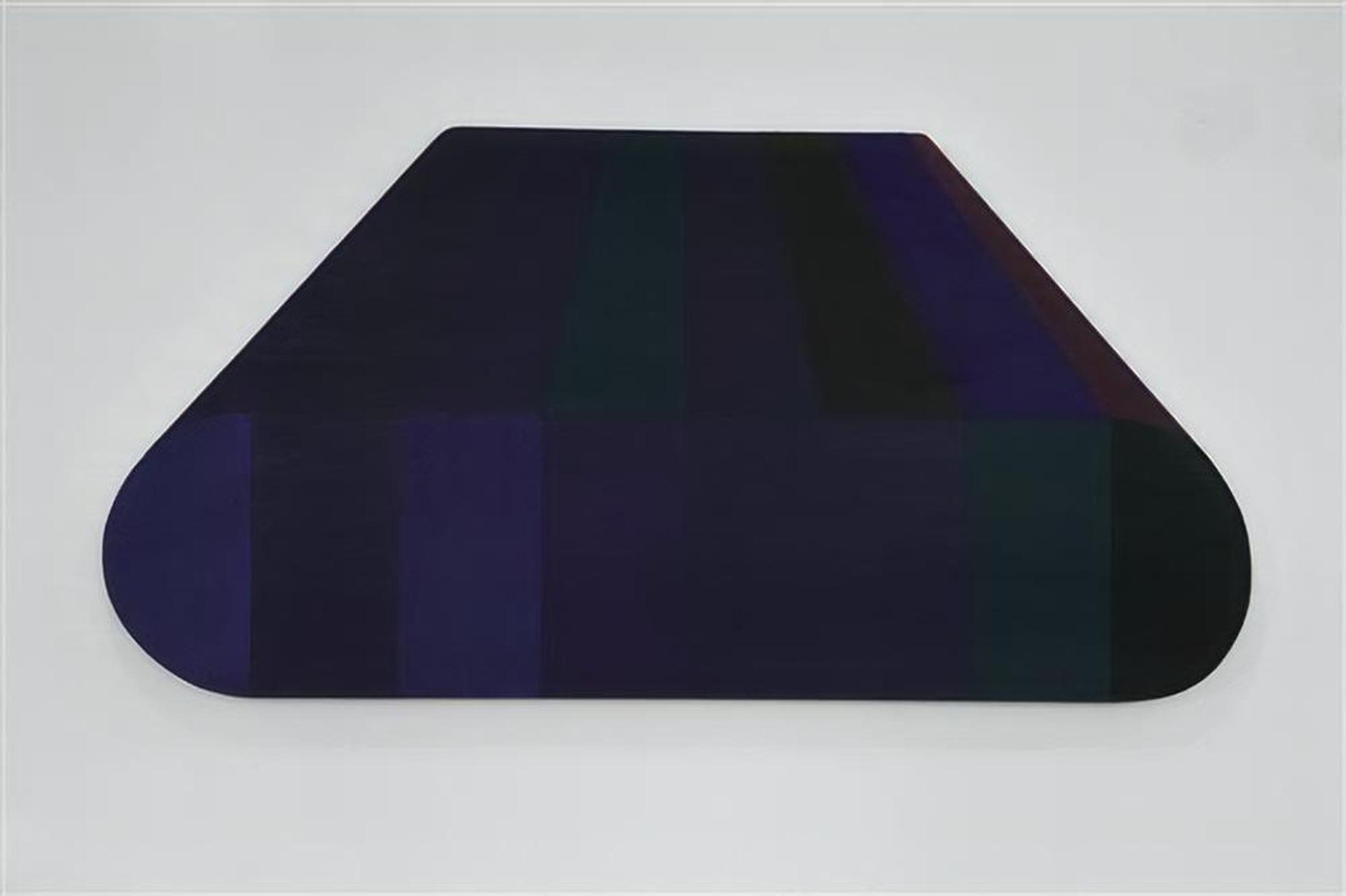}
&\includegraphics[height=0.135\textwidth, width=0.135\textwidth]{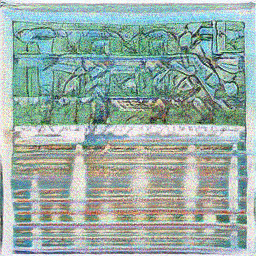}
   &\includegraphics[height=0.135\textwidth, width=0.135\textwidth]{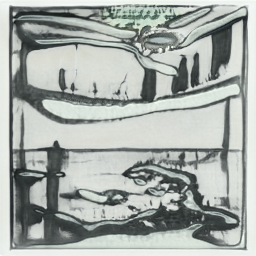}

    &\includegraphics[height=0.135\textwidth, width=0.135\textwidth]{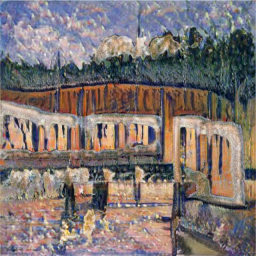}
     &\includegraphics[height=0.135\textwidth, width=0.135\textwidth]{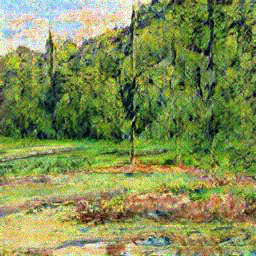}
     &\includegraphics[height=0.135\textwidth, width=0.135\textwidth]{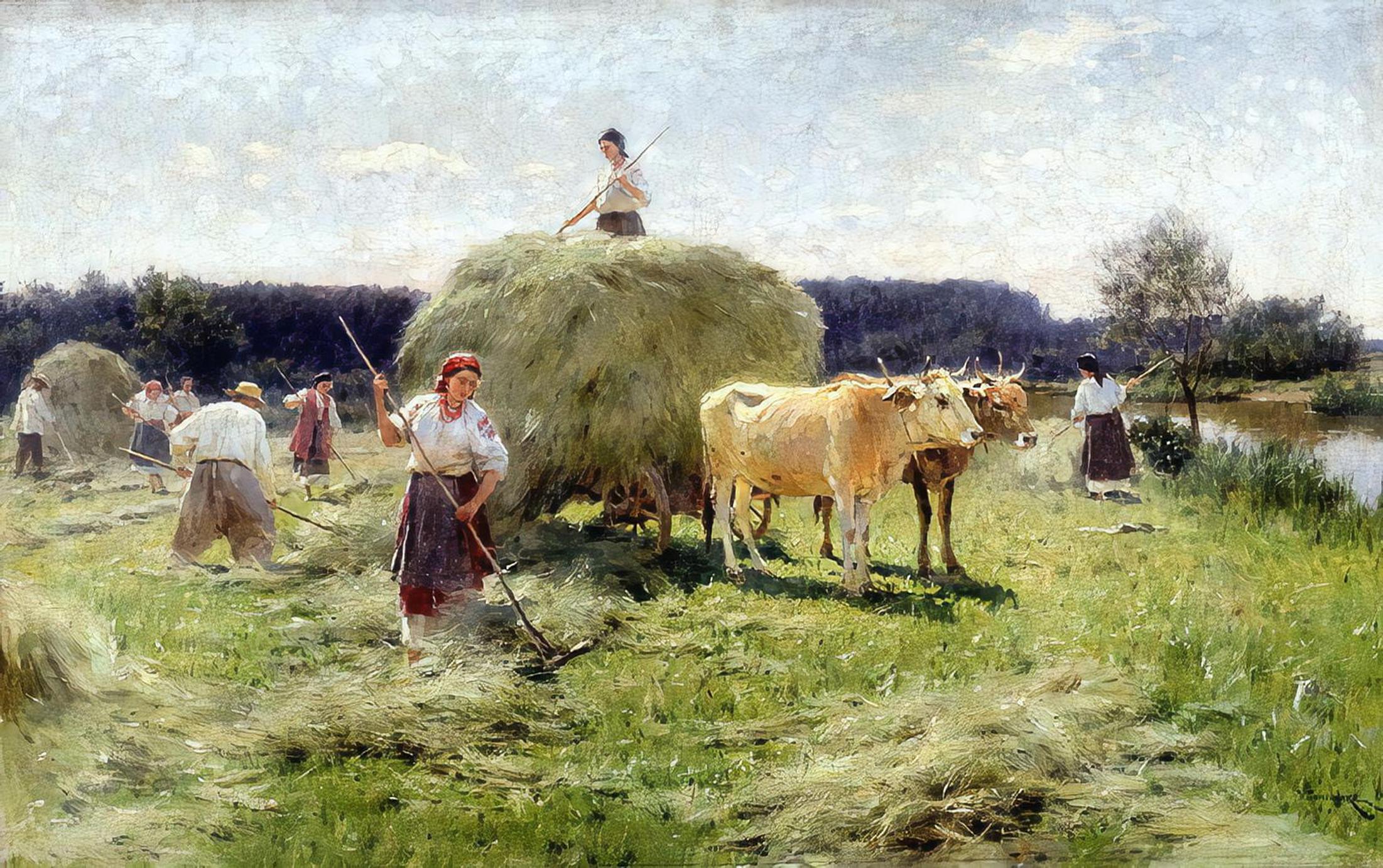}\\
(scores)&($0.69,0.0$)&($0.69,0.0$)&($0.69,0.0$)&($7.19,-3.17$)&($7.38,-3.23$)&($7.48,-3.26$)\\\midrule

rough&\includegraphics[height=0.135\textwidth, width=0.135\textwidth]{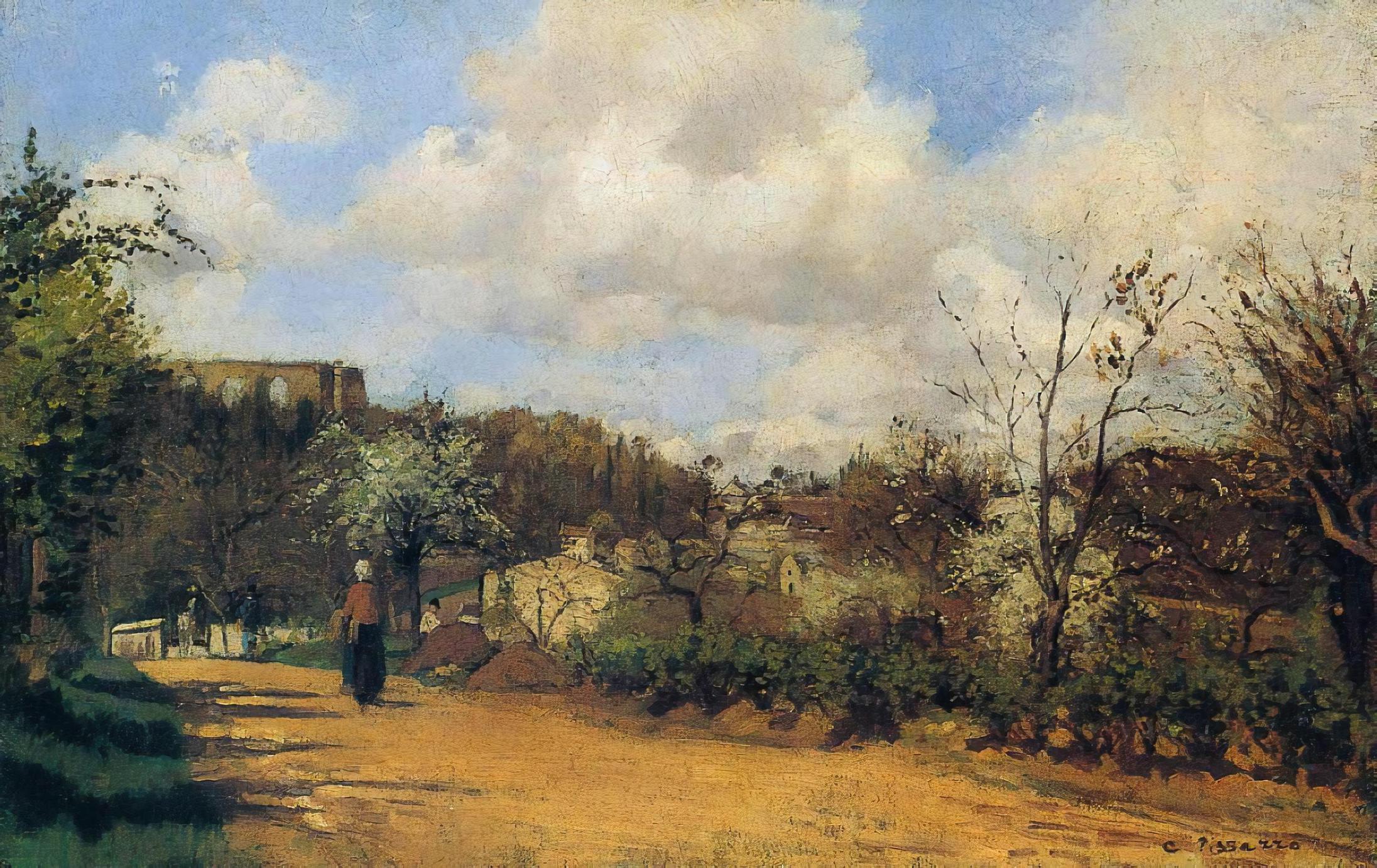}
&\includegraphics[height=0.135\textwidth, width=0.135\textwidth]{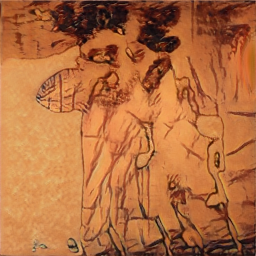}
   &\includegraphics[height=0.135\textwidth, width=0.135\textwidth]{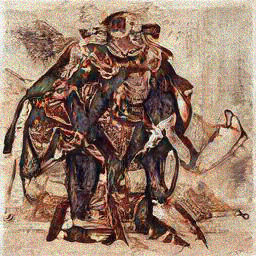}
   &\includegraphics[height=0.135\textwidth, width=0.135\textwidth]{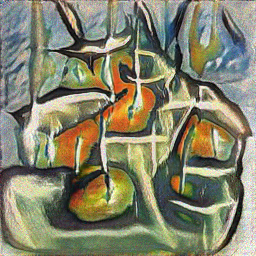}
     &\includegraphics[height=0.135\textwidth, width=0.135\textwidth]{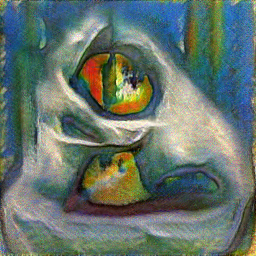}
     &\includegraphics[height=0.135\textwidth, width=0.135\textwidth]{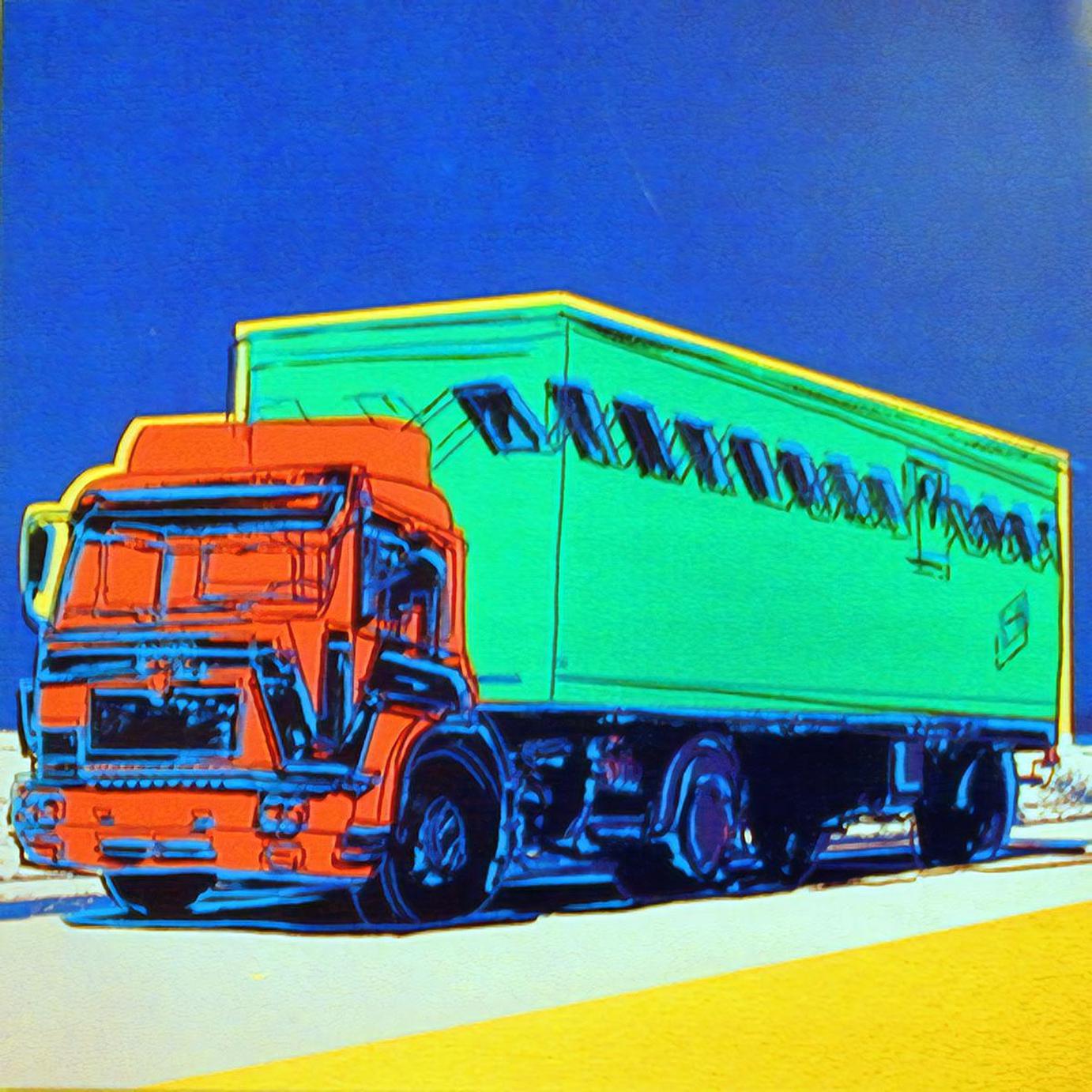}\\
   (scores)&($0.69,0.0$)&($0.69,0.0$)&($0.69,0.0$)&($7.44,3.24$)&($7.95,3.39$)&($8.60,3.56$)\\\midrule

perspective&\includegraphics[height=0.135\textwidth, width=0.135\textwidth]{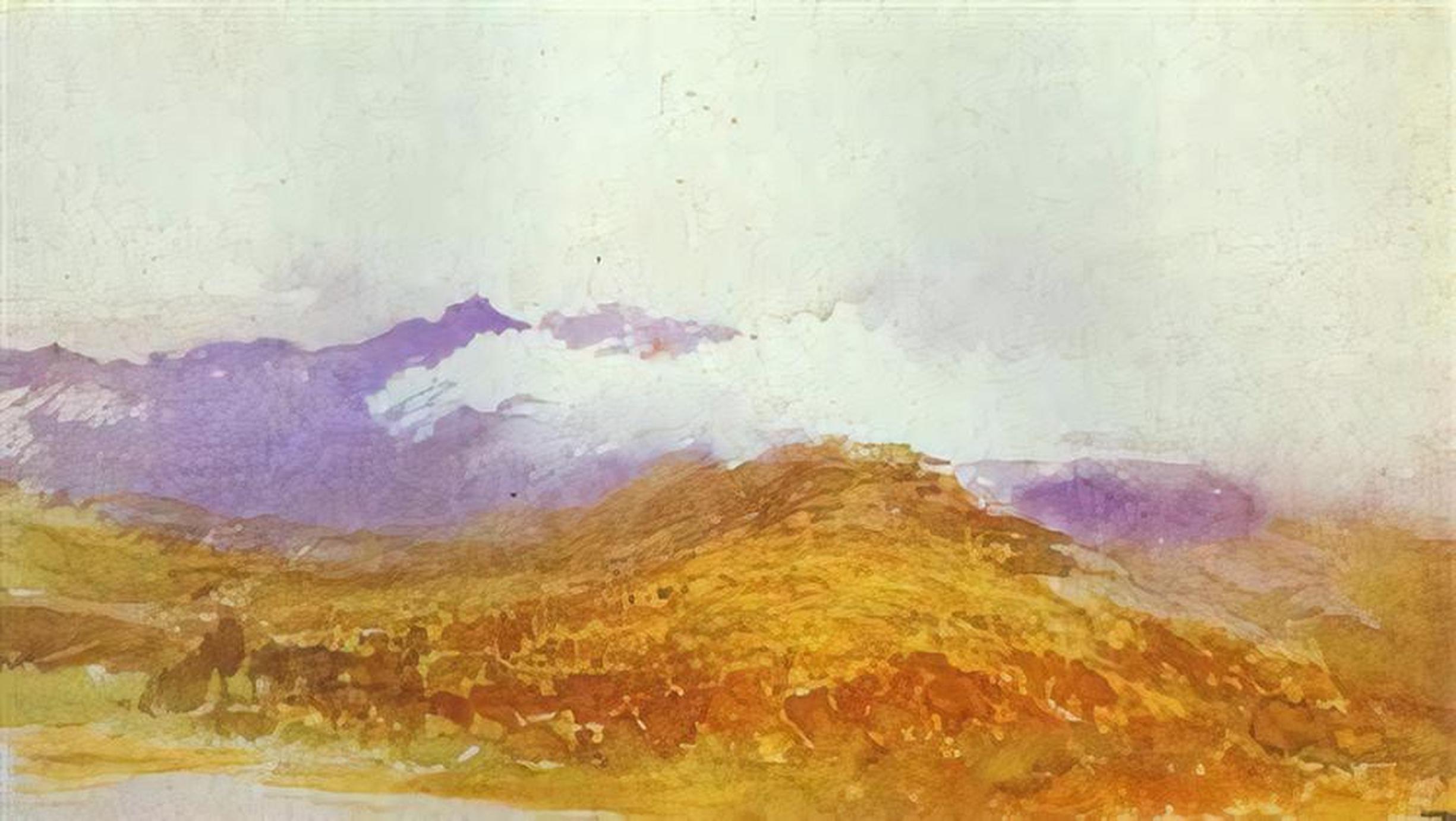}
&\includegraphics[height=0.135\textwidth, width=0.135\textwidth]{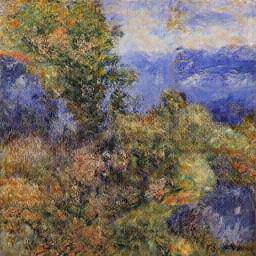}
   &\includegraphics[height=0.135\textwidth, width=0.135\textwidth]{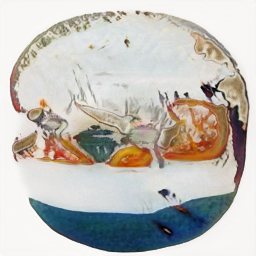}
    &\includegraphics[height=0.135\textwidth, width=0.135\textwidth]{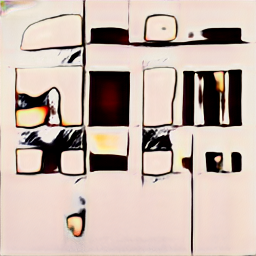}
     &\includegraphics[height=0.135\textwidth, width=0.135\textwidth]{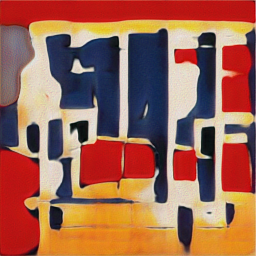}
     &\includegraphics[height=0.135\textwidth, width=0.135\textwidth]{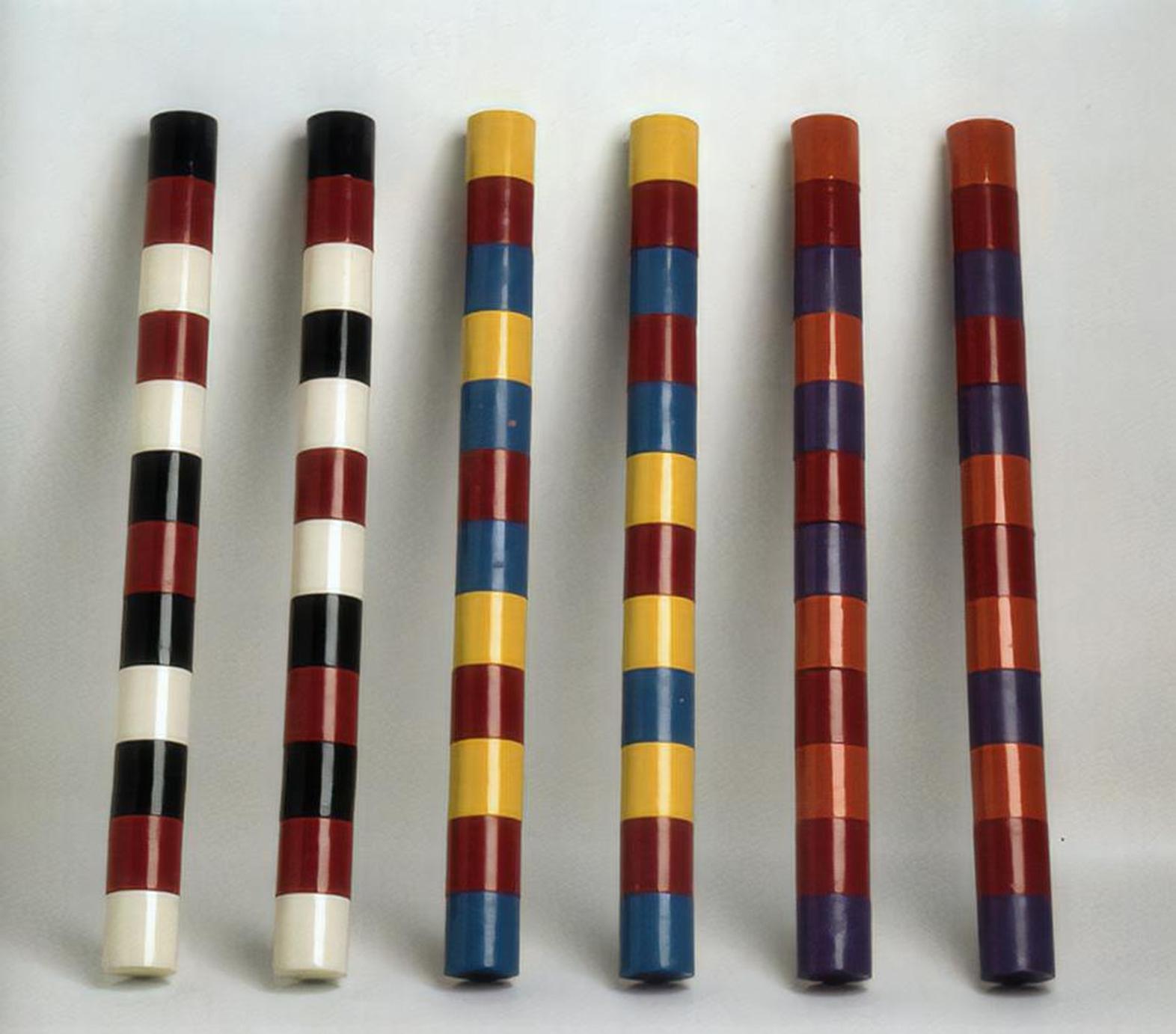}\\
 (scores)&($0.69,0.0$)&($0.69,0.0$)&($0.69,0.0$)&($8.69,-3.59)$&($9.60,-3.82$)&($13.0,-4.58$)\\\midrule

distorted&\includegraphics[height=0.135\textwidth, width=0.135\textwidth]{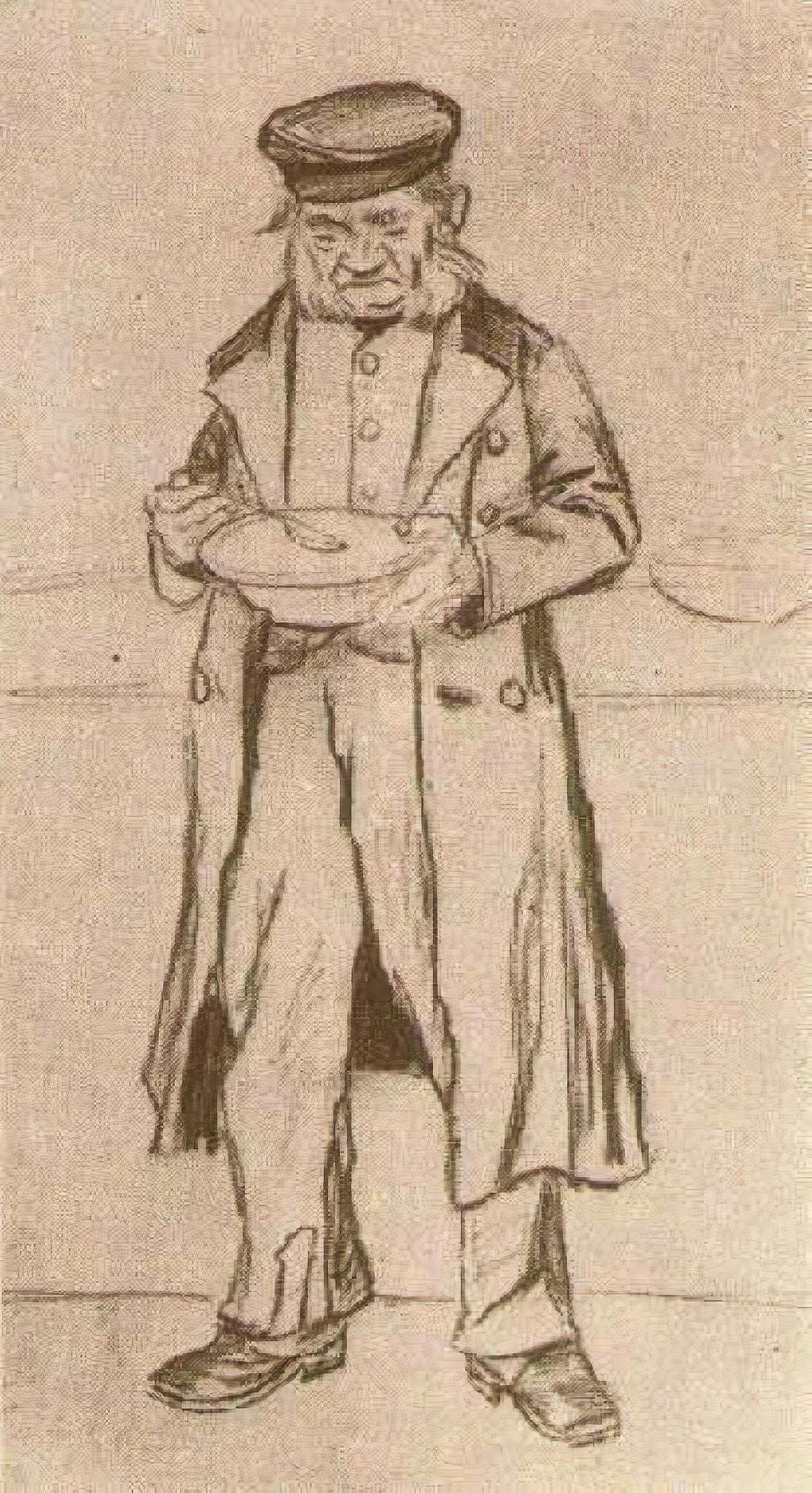}
&\includegraphics[height=0.135\textwidth, width=0.135\textwidth]{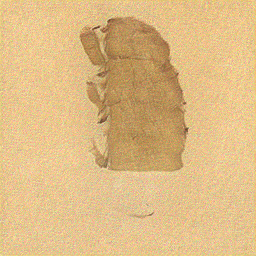}
   &\includegraphics[height=0.135\textwidth, width=0.135\textwidth]{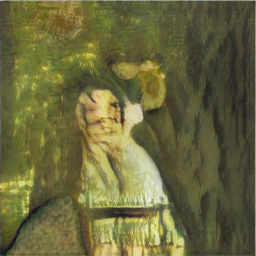}

    &\includegraphics[height=0.135\textwidth, width=0.135\textwidth]{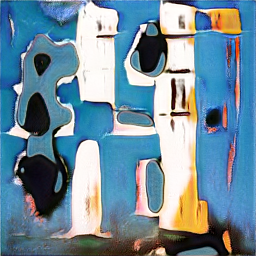}
     &\includegraphics[height=0.135\textwidth, width=0.135\textwidth]{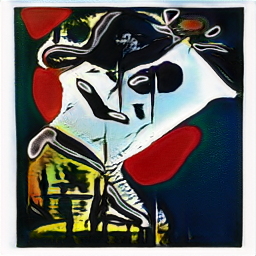}
     &\includegraphics[height=0.135\textwidth, width=0.135\textwidth]{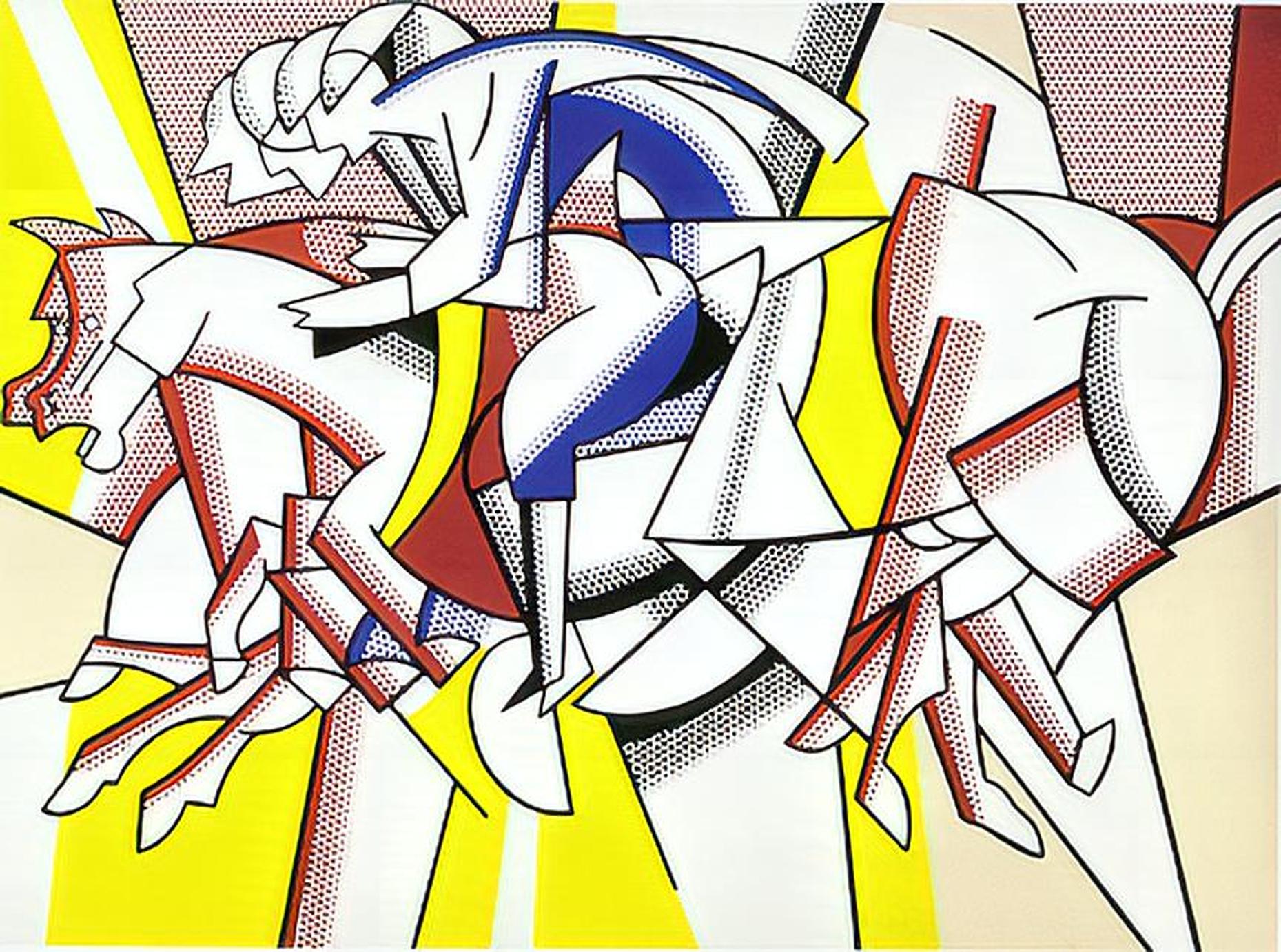}\\
   (scores)&($0.69,0.0$)&($0.69,0.0$)&($0.69,0.0$)&($7.97,3.39$)&($9.52,3.80$)&($16.46,5.26$)\\\midrule

       ambiguous&\includegraphics[height=0.135\textwidth, width=0.135\textwidth]{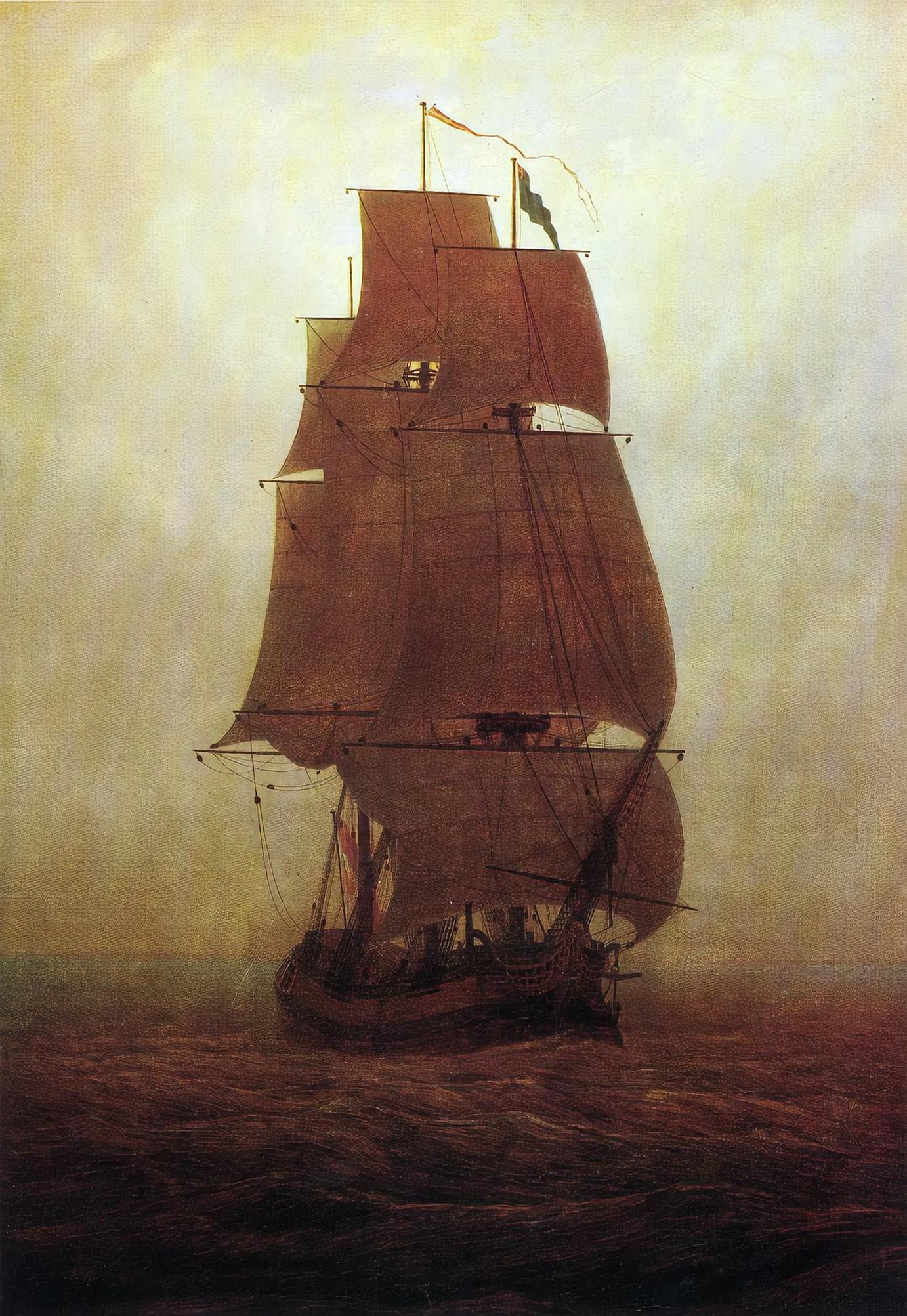}
&\includegraphics[height=0.135\textwidth, width=0.135\textwidth]{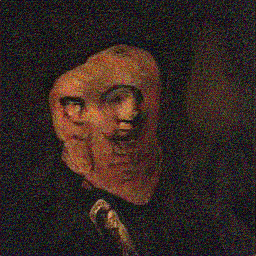}
   &\includegraphics[height=0.135\textwidth, width=0.135\textwidth]{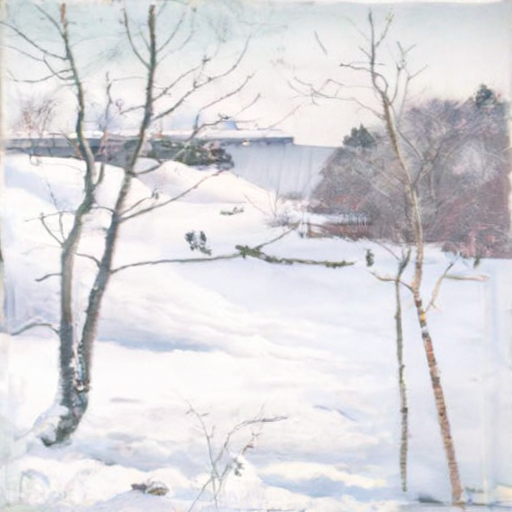}
 
    &\includegraphics[height=0.135\textwidth, width=0.135\textwidth]{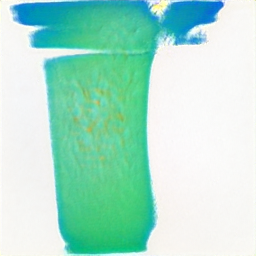}
     &\includegraphics[height=0.135\textwidth, width=0.135\textwidth]{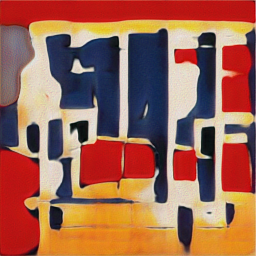}
     &\includegraphics[height=0.135\textwidth, width=0.135\textwidth]{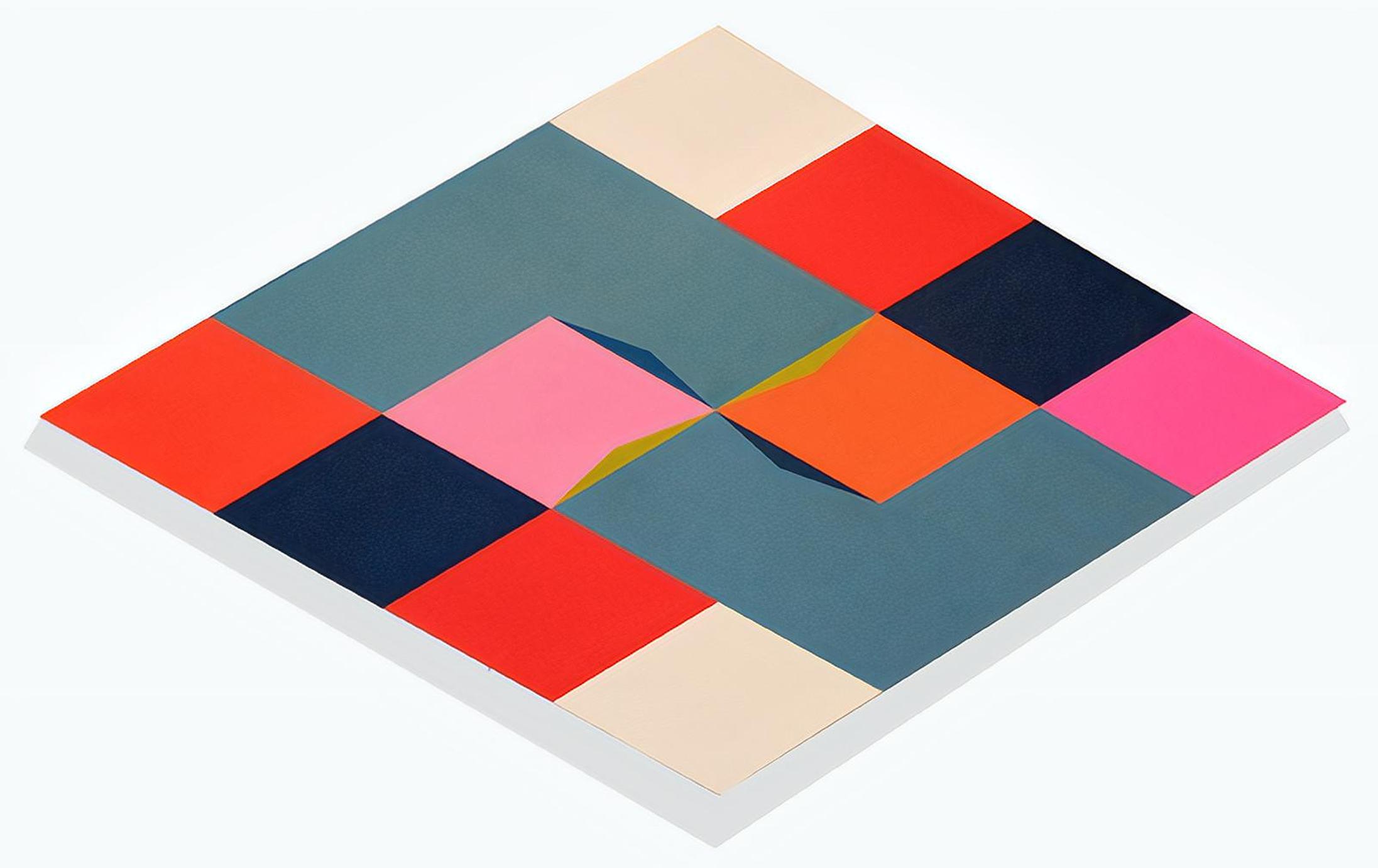}\\
   (scores)&($0.69,0.0$)&($0.69, 0.0$)&($0.69, 0.0$)&($10.29, 3.98$)&($10.66, 4.07$)&($17.85, 5.51$)\\\midrule

\textcolor{red}{mean over 15 concepts}&\includegraphics[height=0.135\textwidth, width=0.135\textwidth]{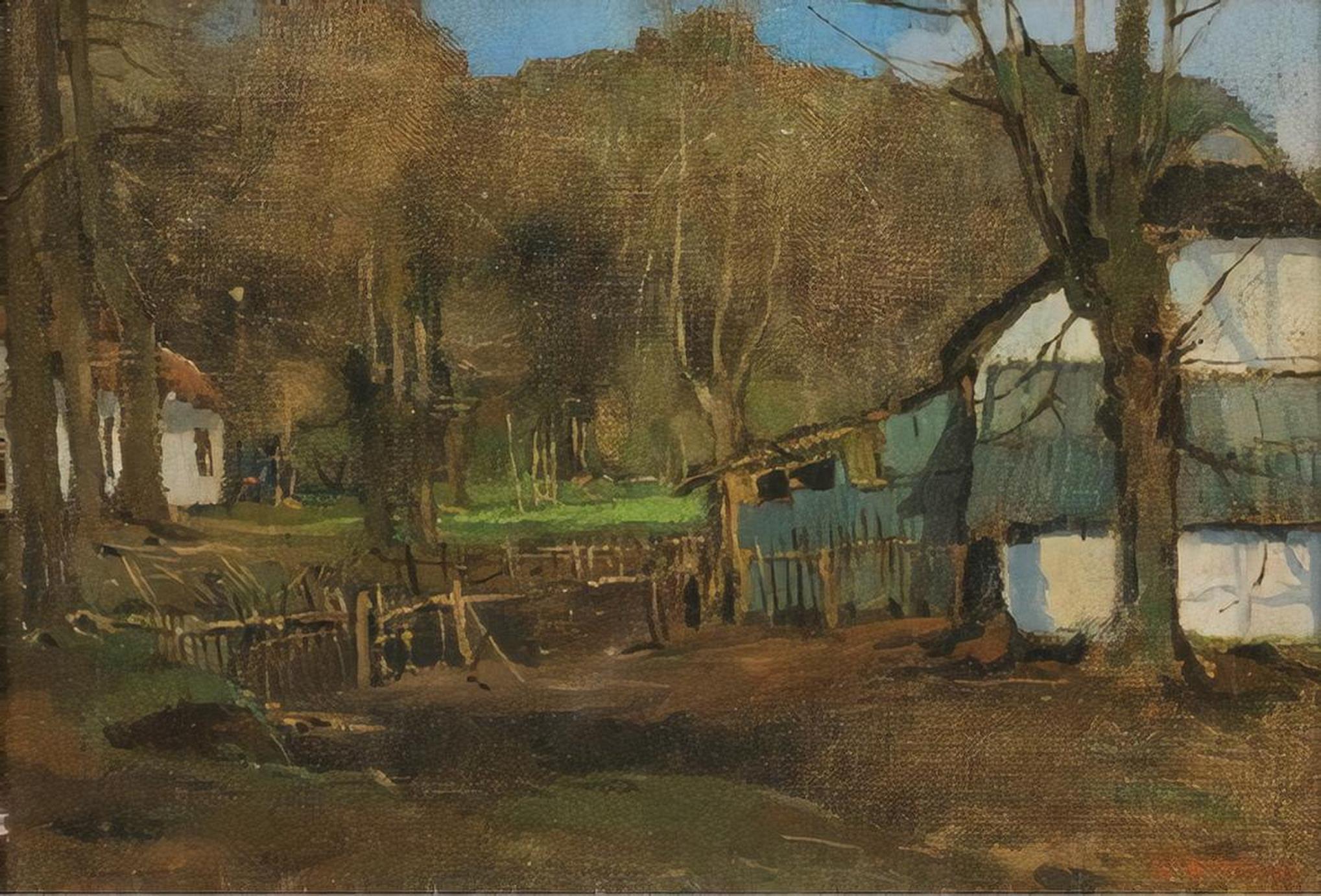}
&\includegraphics[height=0.135\textwidth, width=0.135\textwidth]{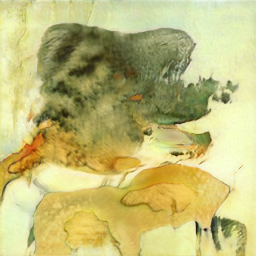}
   &\includegraphics[height=0.135\textwidth, width=0.135\textwidth]{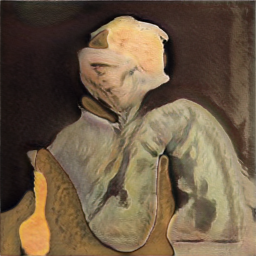}
 
    &\includegraphics[height=0.135\textwidth, width=0.135\textwidth]{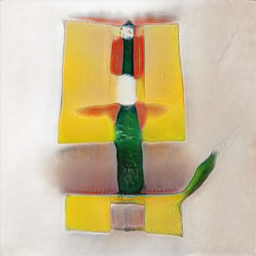}
     &\includegraphics[height=0.135\textwidth, width=0.135\textwidth]{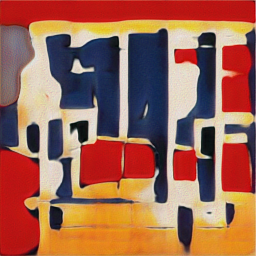}
     &\includegraphics[height=0.135\textwidth, width=0.135\textwidth]{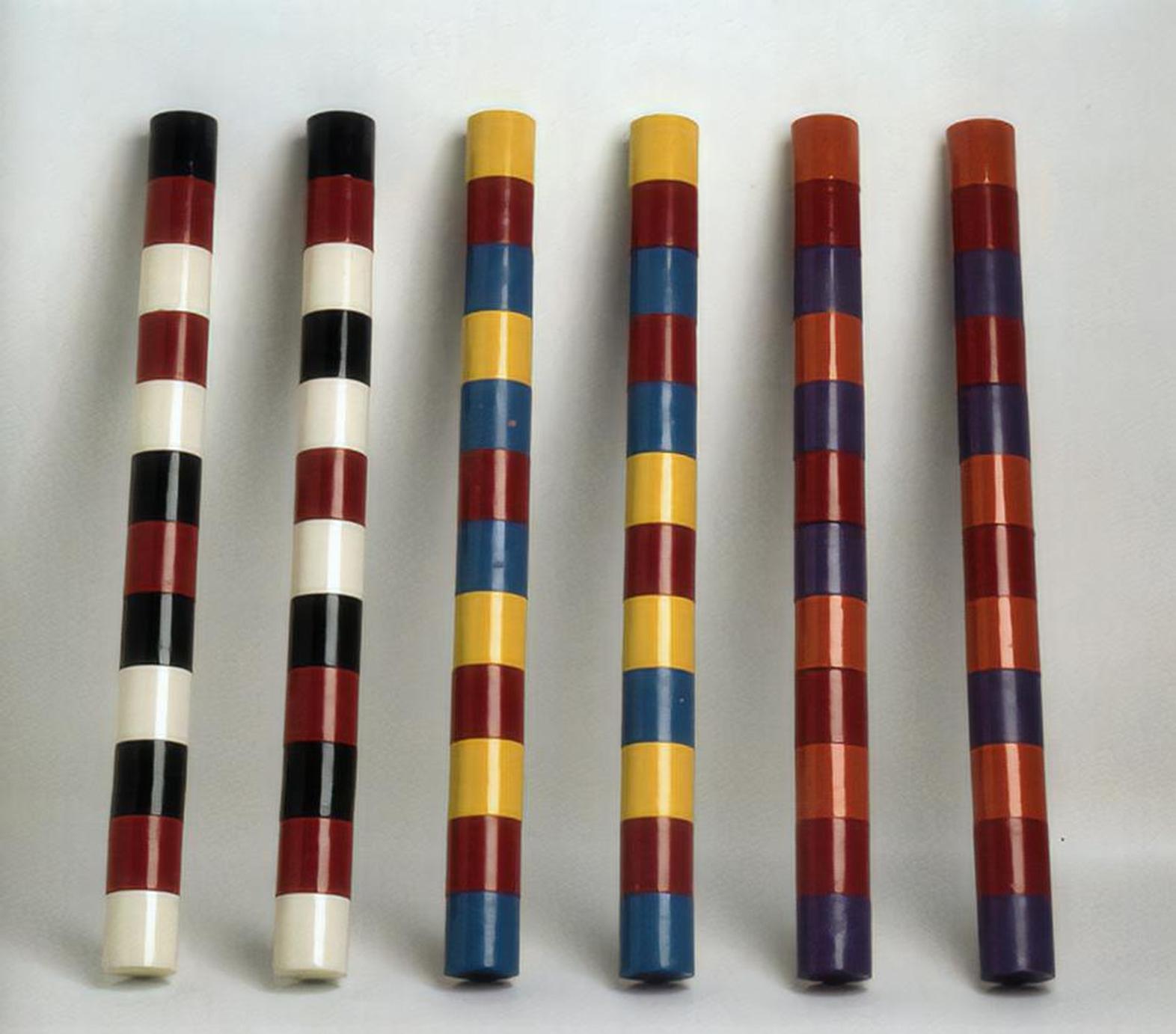}\\
(OODness-proxy only)&($0.66,-$)&($0.70.-$)&($0.71,-$)&($6.47,-$)&($7.34,-$)&($10.60,-$)\\\midrule

\end{tabular}
\end{adjustbox}
\caption{Based on standardized values by human art, the most insignificant (columns 2 and 3) and significant (columns 4 and 5) AI-generated artworks are presented with the most insignificant (column 1) and significant (column 6) human art with the two scores in brackets: OODness-proxy and normalized values. Significant machine samples are observed on the same side of the bias toward modern periods, as seen in our hypothesis test, while still less valued than the most out-lied human art for all 15 concepts. The last row samples are selected by the mean of OODness-proxy of 15 visual concepts.} 
\label{tab:proxy-id-ood}
\vspace{-4pt}
\end{table*}

In Table~\ref{tab:proxy-id-ood}, the most insignificant (columns 2 and 3) and significant (columns 4 and 5) generated arts are presented with the most insignificant (column 1) and significant (column 6) human art with the two scores: (OODness-proxy,  normalized values) for example concepts. For the insignificant sample in the three left columns, we observed AI-generated samples trying to represent the typical subjects in human art: portraits and landscapes, but not in delicate or detailed expressions as much as in human art. 

Another thing to note is that for all visual concepts, significant generated samples are observed on the same side of the bias toward modern arts, as seen in Fig.~\ref{fig:density}. However, we found that the AI-generated art samples are not valued beyond the extremes of human samples in both negative and positive sides. This indicates that generated arts are within the distribution of human arts at least from the aspects of the concepts of proxy-space. The last row samples in Table~\ref{tab:proxy-id-ood} are sorted based on the average of OODness-proxy over all 15 visual concepts to find generated samples that are closest to human art or furthest by art principles. The 2nd and 3rd paintings like the underpainting of a portrait are found to be the closest ones for human art, but the 4th and 5th abstract paintings are the most distant ones. Again their normalized values are less than the most out-lied human artwork: ``Six Round Wooden Bars" by Andre Cadere (1975).

\begin{figure*}[t]
    \centering
    \includegraphics[width=1.0\linewidth]{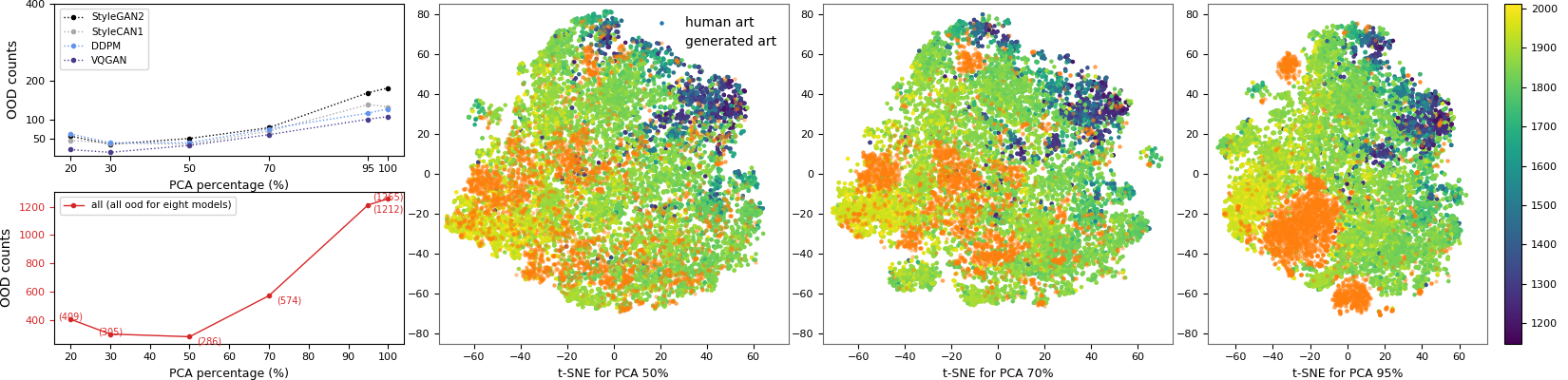}
    \caption{For the various CLIP spaces by different PCA dimensional reductions,  OOD states are compared along with t-SNE representations (perplexity 30): human (greenish dots) and generated arts (orange dots). We found that for the top principal axes from 1 to 17 (50\% data variance), OOD is relatively few, but the number of OOD is increased as more additional marginal dimensions are considered. This phenomenon is observed in both OOD counts and t-SNE plots. }
    \label{fig:ood-tsne}
\end{figure*}

\begin{figure*}[t]
    \centering
    \includegraphics[width=1.0\linewidth]{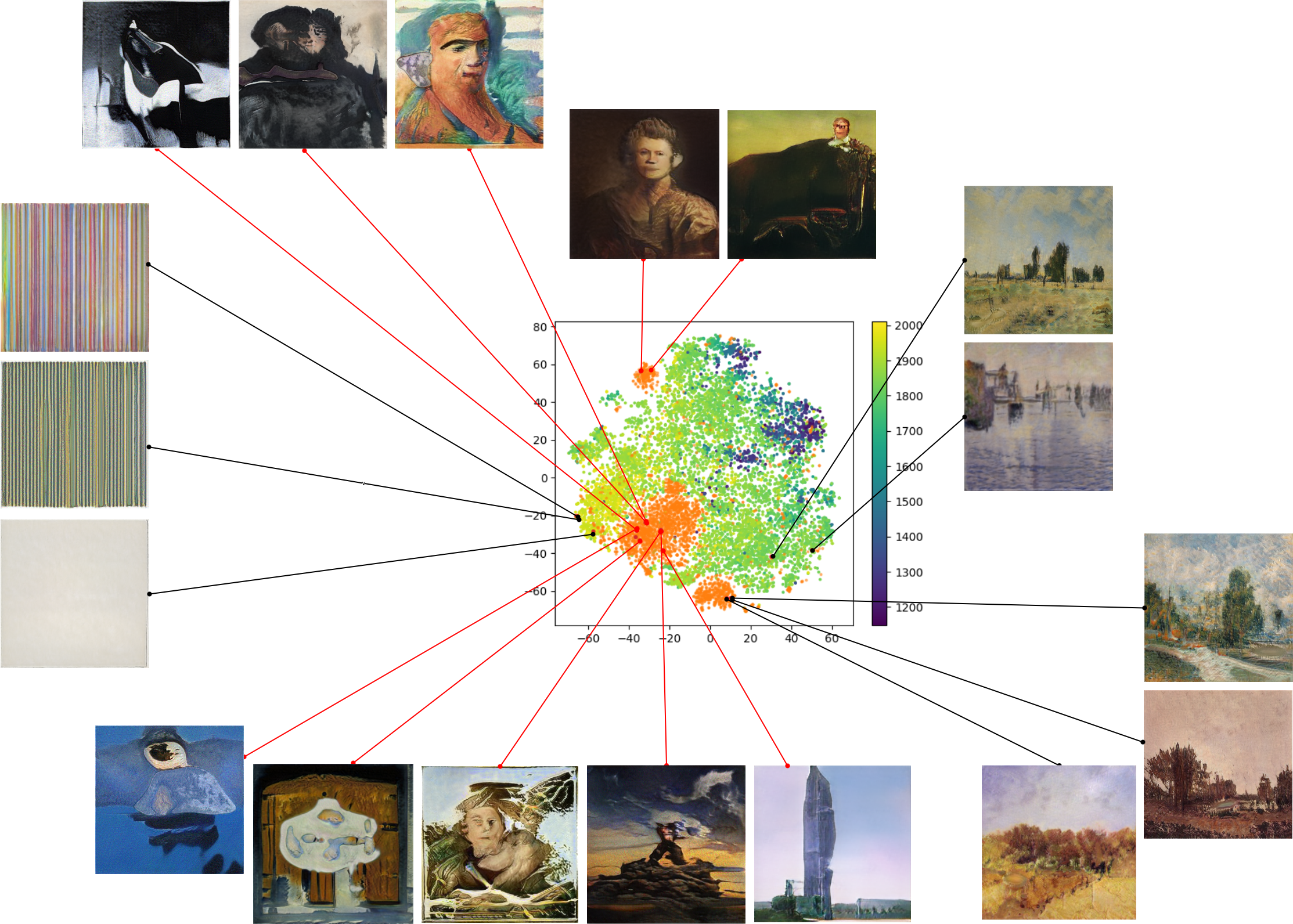}
    \caption{The most ten OOD (red lines) and eight ID samples (black lines) in 95\% PCA are presented along with their t-SNE locations. All ID samples show landscape or abstract and geometric patterns, but OOD samples show irregular and twisted figures.}
    \label{fig:ood-tsne95-painting}
\end{figure*}

\begin{figure}[t]
    \centering
    \includegraphics[width=1.0\linewidth]{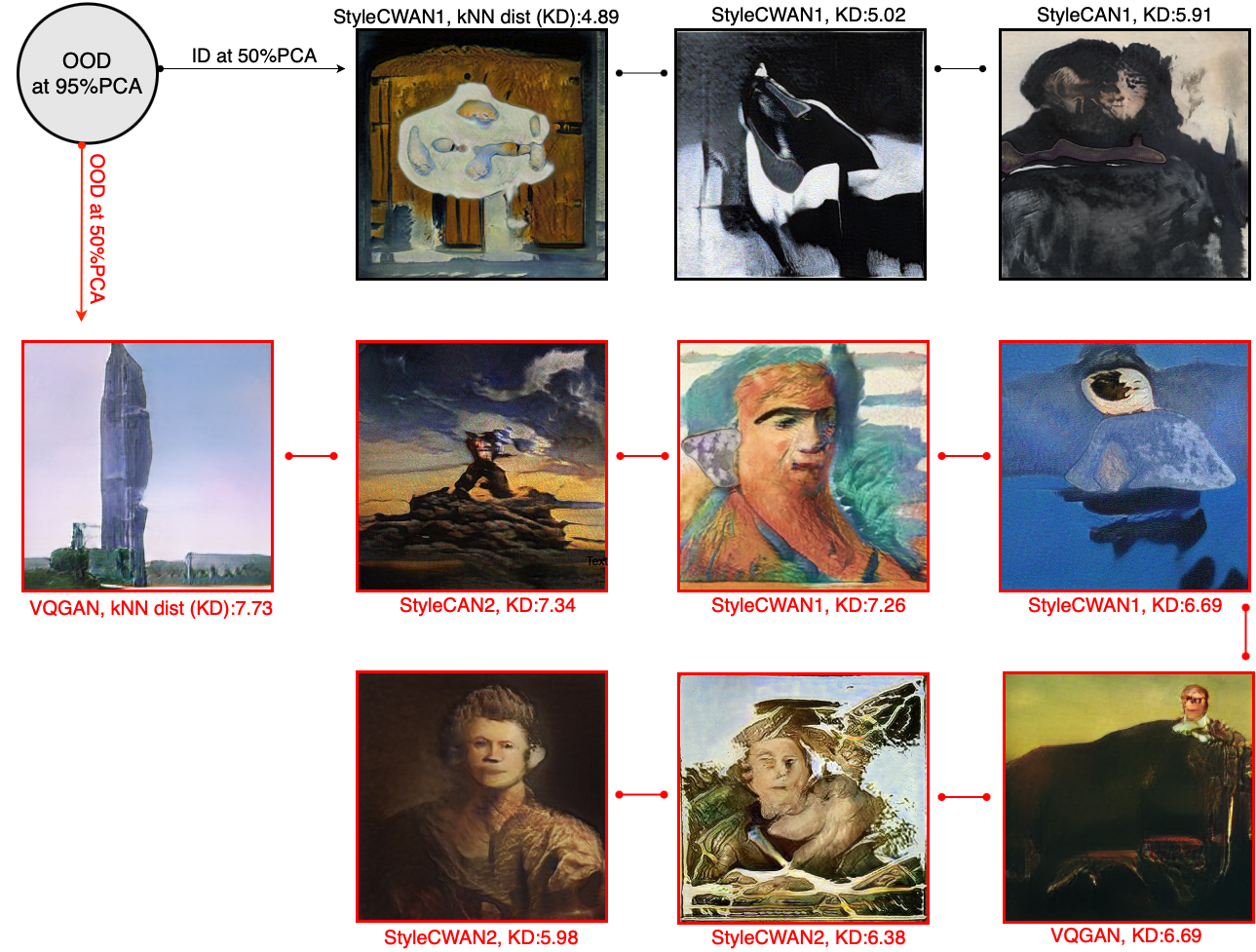}
    \caption{OODness in 50\% PCA space is rechecked for the AI-generated samples detected as most OOD in 95\% PCA space. The three abstract paintings in the first row are newly detected as ID, while the other generated images in the second and third rows are still OOD. The incomplete and unbalanced figuration in the seven samples is what is hardly seen in human art in general.}
    \label{fig:ood_id_split}
\end{figure}

\subsection{Out-Of-Distribution (OOD) Analysis on CLIP}
Motivated by the strong capacity of CLIP \cite{clip_radford} to represent fine-grained visual concepts in images, we conducted an  OOD analysis between human and generated art to see their visual difference in the CLIP space. From the very initial trial, surprisingly, human and generated samples were separated well in CLIP; a third of generated samples are recognized as OOD to human art at 95\% PCA space. 

To investigate the space more, for different PCA spaces: 20\%, 30\%, 50\%, 70\%, 95\%, 100\%, generated OOD samples are detected for each space after computing two kNN distance populations: first is between human vs. human, and the second is AI-generated vs. human~\cite{sun2022knnood}. Then, generated samples are detected as OOD if their kNN distances are larger than a threshold; the threshold is set to the point whose false OOD probability is 5\% (detecting human samples as OOD). All steps followed the OOD framework~\cite{sun2022knnood}, but we did not do normalization in distance computation to use PCA for noise reduction. Fig.~\ref{fig:ood-tsne} shows how OOD counts are varied and how the spatial arrangements among generated and human arts are changed with PCA. For OOD counts, all models showed similar patterns: in the lower dimensions between  20\% $-$ 50\% PCA, the least number of OOD is observed, but OOD is increased as having more principal axes after that, and the result is directly reflected in the t-SNE plots for 50\%, 70\%, and 95\% PCA. This implies that the top principal axes encode the semantics that are shared among generated and human arts in general, while the arts get differentiated by the later PCA axes.

Different principal axes in CLIP encode different visual semantics, so the spatial arrangements among arts were varied with the semantics with PCA spaces. Even though the semantics implied by each axis are unknown, we know that the larger PCA space will contain more profound semantics capable of discerning visual arts. Hence, we questioned (1) what generated samples are still close to human art when they are encoded and compared by the various semantics in the large PCA space (95\% variance: 1$-$256 axes)? and (2) what generated samples are distant to human art when they are compared by a few but essential semantics consisting of the top PCA space (50\% variance: 1$-$17 axes)? By answering the question, we could qualitatively characterize the features shared between human and AI-generated art, and also the distinctive features machine art uniquely has. 

To check the generated samples closest to human in a large PCA space, we collect the most ID samples in 95\% PCA first and also collect OOD samples in the same space to see if the samples are still detected as OOD even when they are represented by the lower dimensional space like 50\% PCA; the most ID and OOD samples are collected for 95\% PCA and traced how their OODness is changed in  50\% PCA. Fig.~\ref{fig:ood-tsne95-painting} shows the ten OOD (red lines) and eight ID-generated samples (black lines) on the t-SNE plot in 95\% PCA. All ID samples present landscapes or abstract patterns, and OOD samples show irregular and somewhat twisted figures. In their OOD detection by 50\% PCA, all ID samples were ID again, while some OOD samples were changed to ID in the new space. Fig.~\ref{fig:ood_id_split} shows the change in OOD detection by PCA from 95\% to 50\%: the ten OOD samples in 95\% PCA become the three ID and seven OOD samples in 50\% PCA. We noticed visual distinctions between the two divisions. The first group—OOD in 95\%  but ID in 50\%—is too abstract to recognize any figurative objects from the images. In the second group, however, we were able to see some figures such as a tower, portrait, and landscapes even though they are incomplete and malformed. 

The features observed in the sub-samples were consistent for all other OOD and ID samples. In summary, all ID-generated images in 95\% PCA depict one of the subjects: landscape, portrait, and geometric abstract. All OOD-generated images in 50\% PCA contain recognizable objects, but they are somewhat unbalanced, abnormal, and incomplete. Based on the observation, we confirmed two things: (1) human and generated art have subjects in common, but (2) some generated artworks get visually distinct from human art when they try to reproduce the subjects but fail to have the delicacy and variants as much as human art. It generates incomplete and reduced forms but presents a new artistic style. A thing to note in this experiment is that most OOD samples were all from GAN-based models, but any DDPM sample was not observed in most OOD, at least within the highest 50 OOD samples. Instead, some DDPM samples were observed in the 50 ID samples for their resemblance with human art. 

\begin{figure}[t]
\centering
\includegraphics[width=0.9\linewidth]{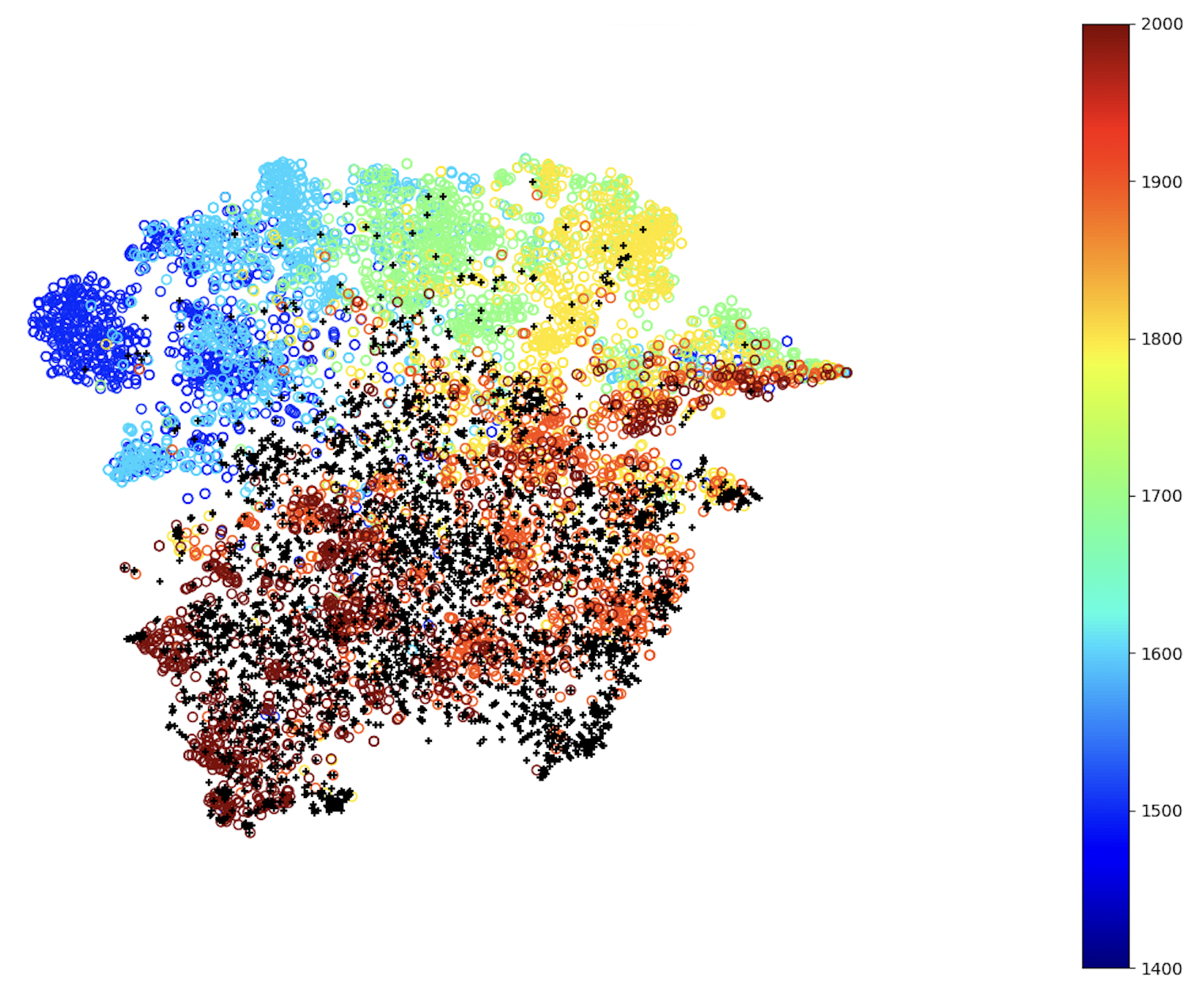}
\caption{For ResNet50, the t-SNE visualizations of features of AI art are overlayed with human art. The black markers are the AI art samples.}
\label{resnet-time}
\end{figure}

\begin{figure}[t]
\centering
\includegraphics[width=0.99\linewidth]{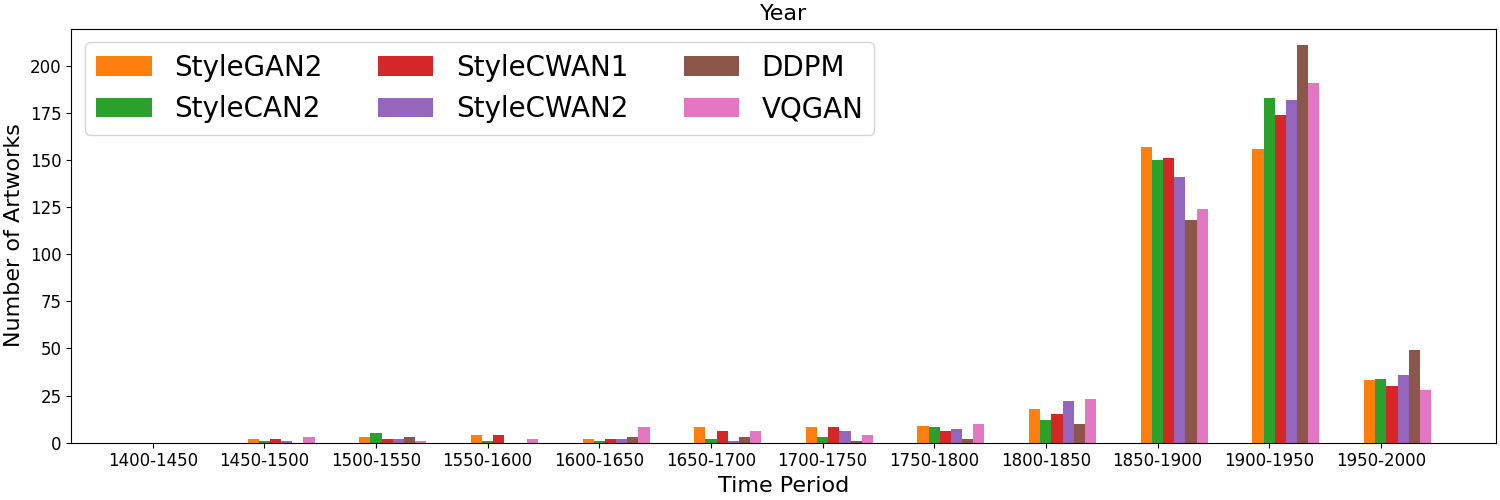}
\caption{The bar plot indicates the number of AI art pieces in their respective periods. The year for the AI artwork is estimated by averaging the nearest neighbor human art's period in  ResNet50.}
\label{resnet-bar}
\end{figure}

\begin{figure*}[t]
\centering
\includegraphics[width=1.0\linewidth]{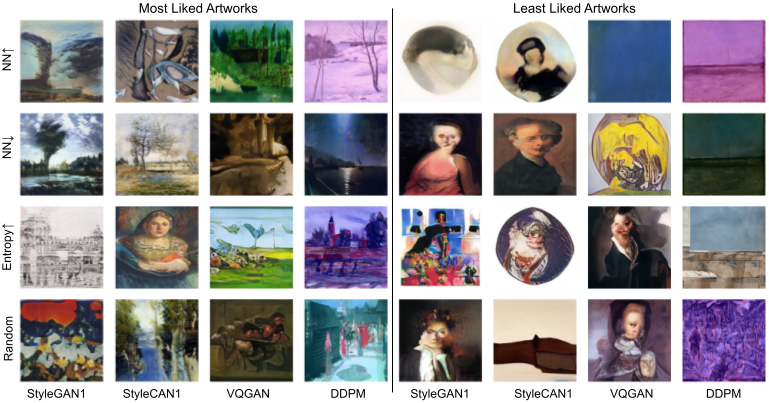}
\caption{The most and least liked AI-generated artworks. There was a constant feature that affected likability throughout all models: the highly likable images are mostly the typical landscape composed of horizontally lined blue sky, water, and trees whereas the low-scored images are deformed and distorted portraits.}
\label{fig:like_dislike}
\end{figure*}

\subsection{Time Analysis}
The two ResNet50 hidden embeddings for 3,200 AI-generated and 6,000 human art are overlaid after t-SNE visualization in Fig. \ref{resnet-time}.  We observed that black-colored AI-generated images cover a particular section of feature space where modern art is dominant. This t-SNE representation implies that the AI-generated art is visually close to modern period human art, which is consistent with the experimental finding of general art principles in section \ref{general art principles_experiment}: the AI-generated art is visually biased to modern period art.
\par To figure out the specific period of modern art closest to our AI-generated art, we computed the following. We first find the five nearest neighbors from human art in the ResNet50 space for each of the generated artworks and then assign the average year of these five nearest human art pieces of art to the generated artwork. We observed that the average year of human art from the WikiArt dataset was $1869$, but it was 1891 for the generated art. In Fig.~\ref{resnet-bar}, we plotted the year distribution of the generated art from various models. The majority (70\%) of the AI-generated art was estimated to be in the period from $1850$ to $2000$. The average year of generated art was $1882$ for Style\textbf{GAN}-2, $1894$ for StyleCAN-2, $1887$ for StyleCWAN-1, $1893$ for StyleCWAN-2, $1885$ for VQ-GAN, and $1902$ from DDPM.

\begin{table}[t]
\begin{center}
\caption{Turing and likability results are shown across different groups and models.  Notably, StyleCWAN consistently exhibited the higher mean likability in all groups compared to the other Style\textbf{GAN} variants, and the LowestNN group, close to human art in ResNet50, showed the highest likability. Also, the DDPM artworks garnered the highest mean likability across all groups. In the Turing test, more than 50\% of the participants believed that the machine artworks were drawn by human artists. As a reference, \cite{elgammal2017can} reported Turning test rates as $85\%$, $41\%$, $62\%$ for the three human art sets: Abstract Expressionist, Art Basel, and Artists Sets Combined. Some qualitative AI art examples and their Turing results are presented in Appendix~\ref{turingtest}. }
\label{tb:likability}
\begin{adjustbox}{width=0.475\textwidth}
\begin{tabular}
{c|ccccc|c}
\toprule
{}&\multicolumn{5}{c|}{Likability}&
{Truing test}\\\midrule
{models}&\multicolumn{1}{c|}{Q1-mean/std}&{NN$\uparrow$}&{NN$\downarrow$}&{Entropy$\uparrow$}&{Random}&{Q2-human artists (\%)}\\\midrule
Style\textbf{GAN}-1&\multicolumn{1}{c|}{3.12/0.58}&3.07&3.36&3.00&3.06&55.53\\
Style\textbf{GAN}-2&\multicolumn{1}{c|}{3.02/0.67}&2.89&3.31&2.79&3.09&53.80\\
StyleCAN-1&\multicolumn{1}{c|}{3.20/1.14}&3.01&3.61&3.05&3.11&56.55\\
StyleCAN-2&\multicolumn{1}{c|}{3.23/0.61}&3.27&3.34&3.11&3.21&57.70\\
StyleCWAN-1&\multicolumn{1}{c|}{3.29/1.12}&3.15&3.67&3.15&3.17&58.63\\
StyleCWAN-2&\multicolumn{1}{c|}{3.40/1.10}&3.30&3.61&3.33&3.35&64.00\\
VQ-GAN&\multicolumn{1}{c|}{3.57/1.03}&3.55&3.65&3.57&3.52&65.90\\
DDPM&\multicolumn{1}{c|}{3.85/0.91}&3.77&3.90&3.81&3.93&63.55\\\bottomrule
\end{tabular}
\end{adjustbox}
\end{center}
\end{table}

\subsection{Beholder's Visual Experience}
Likability and emotion are the last elements for the comparison of the artworks. We conducted a human survey and found that AI-generated artworks are more likable when they successfully depict the landscape and portrait, typical genres in human art,  and they evoke a diverse range of emotions, including anger, awe, contentment, disgust, excitement, fear, sadness, and amusement. Participants identified various visual features that contributed to the likability of generated art, such as unique color combinations, imaginative compositions, intricate details, and captivating narratives. These findings highlight the potential of machine generative art in offering engaging and emotional impact on artistic experiences.

\subsubsection{Likability}
For all 3,200 AI-generated artworks, we conducted a human survey for likability. The participants are asked two questions: Q1 ``How much do you like this image?"  and   Q2 `'Whether the art is created by a human artist or a machine?'' The rating of Q1 is ranged from 1 (extremely dislike), 2 (like), 3 (neutral), 4 (like), and 5 (extremely like), and Q2 is a binary test. Table~\ref{tb:likability} shows the numerical results: the mean values for the first question (likability) and the percentage of the answers of human artists  (Turing Test). For the mean score, all machine artworks are scored higher than 3: neutral, but less than 4: like. In the Turing test,  more than half of the subjects thought our AI-generated samples were created by human artists. For qualitative analysis, in Fig.~\ref{fig:like_dislike}, the most and least liked machine art samples are presented according to the four groups of the section \ref{group}.
\par We have three points to remark from the figure and table. First, StyleCAN-1\&2 and StyleCWAN-1\&2 scored higher than Style\textbf{GAN}-1\&2; in the survey, the models trained with creative losses resulted in 38\% and 18\% more people assigning a full score of 5 over the vanilla models. Second, DDPM is liked the most (score 3.81) than any other model. DDPM images are all figurative and representational. We observed some samples quite resemble specific paintings by human artists. For example, the most liked HighestNN in Fig. \ref{fig:like_dislike} reminds us of the winter landscapes of Alfred Sisley. We also found multiple DDPM variations of Marcel Duchamp's \textit{Sad Young Man on a Train} of different tones in color. Lastly, in comparison of the two NN groups (Highest NN vs.  Lowest NN), Lowest NN—visually closer to WikiArt—got higher likable scores than other groups for all the generative models. In our empirical observation, we found a correlation between the closeness to human art and likability. AI-generated artworks were more likable when they successfully depict the popular and familiar topics of landscape and portrait in human art, but they are disliked much when they fail to be drawn as a sound shape, like the deformed portraits in Fig. \ref{fig:like_dislike}. According to groups and likability, we characterize our generative models empirically as written below.\\

\noindent\textbf{HighestNN:}
the generated artworks with the highest likability in HighestNN are similar to modern abstract art with contrasting solid colors. Paintings by Style\textbf{GAN}-1 tended to have a more disorderly composition and unidentifiable figures, whereas StyleCAN-1 demonstrated identifiable abstract figures in a more orderly manner. The StyleCAN-2 and StyleCWAN-2 with a high mean likability (rated 4 to 4.4) displayed artistic characteristics reminiscent of Salvador Dali's surreal paintings. The balance of surreal figures and orderly arrangement of the overall composition make the artwork interesting and comforting for the viewer. In contrast, StyleCWAN-1 artwork in this category had architectural characteristics, sometimes with disorderly compositions. Style\textbf{GAN}-2 artwork had architectural characteristics and geometric shapes. The artworks reminded the viewers of Kandinsky's \textit{Composition} painting series. In HighestNN by VQ-GAN, we found many landscape-like abstract paintings to have vivid, thin, and patterned strokes in vertical, horizontal, or sometimes circular directions. It reminds us of Claude Monet's thin, dense strokes and loose swirls often appearing in his late artworks. \\

\noindent\textbf{LowestNN:} this group showed consistent characteristics in terms of style and subject matters in general. The most liked images exhibited an impressionist style marked by vibrant colors and rapid and distinct brush strokes. The subject matter of the liked artworks is mostly natural landscapes. These art pieces exhibited similarities to Claude Monet's iconic impressionist paintings, such as \textit{Impression Sunrise} and  \textit{The Garden of Monet at Argenteuil.} This group is collected for the close nearest neighbor distance from WikiArt in ResNet50, so portrait genres in human art are often shown but disliked by subjects when the figures are contorted. However, for VQ-GAN, LowestNN presents no figuration images, but mostly abstract shapes, and opaque and heavy brush touches are common like the VQ-GAN LowestNN examples in Figure \ref{fig:like_dislike}, it was hard to feel visual speed from the VQ-GAN samples. \\

\noindent\textbf{Highest Shape Entropy:} most paintings in the Highest shape entropy group tend to depict complex compositions and dense arrangements of figures. The complex nature of these paintings may have caused difficulties for the classifier to categorize them. The lively colors and complicated subject matter demonstrate similarities to Hieronymus Bosch's \textit{The Garden Of Earthly Delights}. The clarity of the paintings determined by the distinct depictions of figures affects the mean likability and artist ratings. Paintings with blurred and overly distorted figures tend to receive lower mean likability and artist ratings.\\

\begin{figure}[t]
        \centering
        \includegraphics[width=1\linewidth]{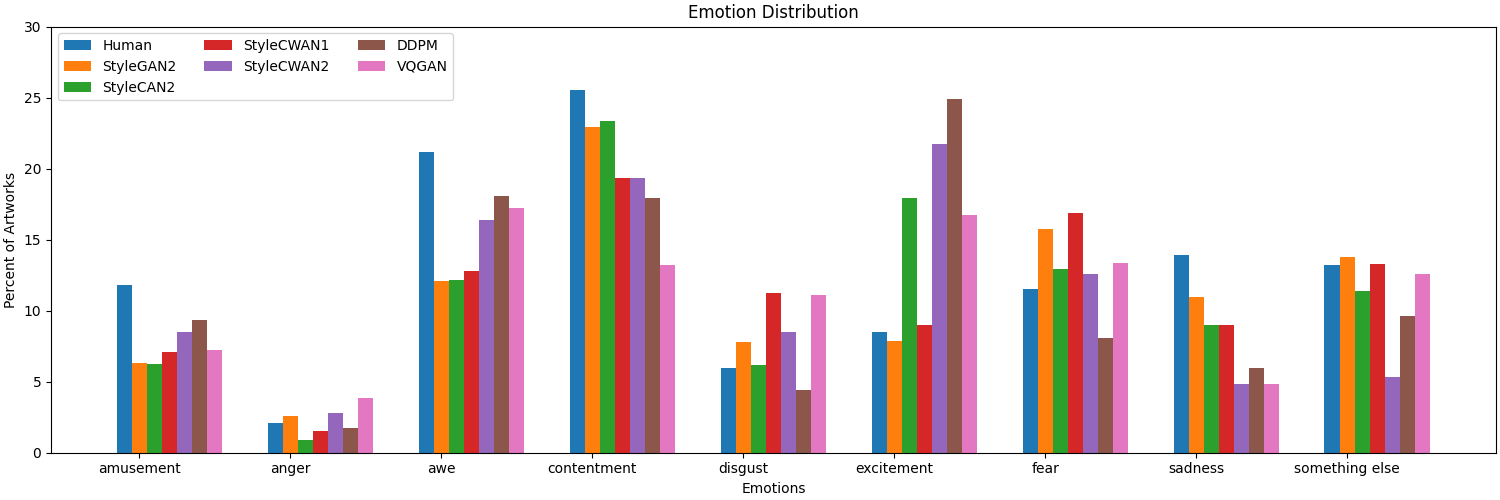}
    \caption{Emotional distribution results of the human survey on human and AI-generated artwork. Both art evoke diverse emotions. Contentment has the highest share, and negative emotions are relatively rare. }
    \label{fig:emo_distri}
\end{figure}

\begin{figure}[t]
        \centering
        \includegraphics[width=1.0\linewidth]{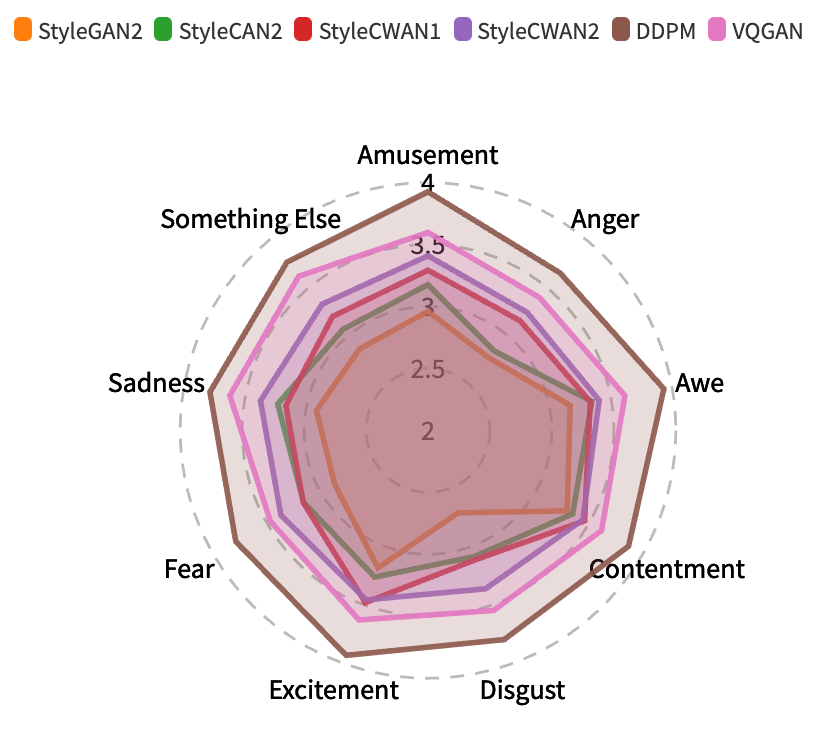}
    \caption{The likability scores for art generated from different models, categorized by emotion. As we computed PCC between likability scores and emotions, we found a weak positive correlation ranging from 0.07 to 0.25. In the analysis, four negative emotions are mapped to -1, four positive emotions are mapped to +1, and “something else” is omitted.}
    \label{fig:all_emo_distri}
\end{figure}
\subsubsection{Emotional Distribution}

In this experiment, we examined the collected data for emotion and investigated the diversity of emotions elicited by AI-generated art. As shown in Fig.~\ref{fig:emo_distri},  AI-generated art is capable of constructing diverse sets of emotions. To measure the diversity of the emotions evoked by generated art, we calculated the entropy of emotions evoked by AI-generated art and compared it with human art. Human art has an entropy of $0.916$, whereas the combined generated art from all models has an entropy of  $0.938$, indicating that generative art is capable of evoking diverse emotions. However, some differences exist in the distribution. For example, excitement accounts for around 17\% of all of the generated art, whereas for human art, it accounts for only 7.47\%. In addition, DDPM has the highest percentage of art evoking excitement of 24.90\%. Awe accounts for 18.63\% of human art, which is much higher than machine art. We also observe that contentment has the highest share of any emotion for both human and generated artworks. We also plot the likability scores per emotion in Fig. \ref{fig:all_emo_distri} for all generated art. \\

\noindent \textbf{Qualitative Analysis of Emotion:} based on the responses of the emotion interface participants, we derived and analyzed common factors and elements of the AI-generated artworks that constructed emotions of anger, awe, contentment, amusement, disgust, excitement, fear, sadness, or something else within the audience. We find that color and composition are the key factors for emotions in the emotional narratives of participants. Also, some familiar and ordinary subjects make people comfortable and contended by bringing a good memory related to them.  Two positive and negative emotions are discussed in detail below.\\

\noindent\textbf{Anger:}
paintings that contain unfamiliar abstract figures tend to evoke confusion and emotions of anger within the audience. As shown in one of the "fear" responses, ``Another kind of painting that makes me ask to myself what does it mean?'' and ``The painting looks confusing and shows no representation.'', some puzzling and ambiguous depictions of machine art caused discomfort and annoyance from participants. The use of colors and artistic composition affects emotional response, too. Some vivid coloration, especially red, was usually associated with the emotion of anger, as stated by the participant, ``The red color used seems like a man bleeding with anger.” We also observed how the color can be associated with the concept of violence; ``This makes me feel anger because the black marks between the rocks look like violent movements." Participants also felt anger in a painting representing a situation of disorder. The disorderly arrangement of figures creates a sense of unease and constructs anger. \\

\noindent\textbf{Awe:}
Artwork that portrays ordinary subject matter, such as a brown coat, a man, a plant, or the sky, tends to construct emotions of awe within the audience, as demonstrated in the participant's statement; ``The plant looks realistic.'' Many participants referred to the use of complementary colors and a soft color scheme as the grounds for their emotions of awe. In addition, artwork depicting natural landscapes also tends to create calming effects and raise awe for natural scenes, as stated by the participants: ``Colors are vibrant.'', ``The play of colors in the sky of this painting is magnificent.",  ``The trees stand tall in the open air and the sky shines brightly.'', and ``Trees are often associated with life, growth, and vitality. In a desert landscape where life may be scarce, the presence of trees can symbolize endurance, strength, and the tenacity of living organisms. This symbolism can evoke a sense of awe as it highlights the ability of nature to thrive in challenging conditions." \\

\noindent\textbf{Contentment:} Floral color arrangement constructed emotions of contentment according to the answers from the participants: ``The light of roses.'',  ``The green color invokes a sense of fertility and contentment.'', or ``Mix of colors between green and yellow reminds the changing of seasons, brings peace and tranquility.'' In addition to color, participants underlined the effect of depth, layers, and orderly composition in their feeling of contentment, or sometimes they associated the paintings with their own experience or memory from the past. These are the example statements: ``This painting makes me feel relaxed because the items are well-ordered and displayed coherently.'', ``The traces of the image remind us of happy everyday things.'', ``I liked the image I felt pleasure because reminder my childhood.'', and ``It reminds me of my grandfather's place where he had ponds and lots of frogs.''\\

\noindent\textbf{Disgust:}
The use of dark color and its visual effects construct emotions of disgust, as stated by the participants: ``Too much darkness on the sea.'', ``Looks like a dark cloud about to eat a human.'', and ``The dark color scheme except for what to me could be a pair of broken glasses makes me feel frustrated.'' In addition, when the artwork did not convey a message or meaning, the audience felt a lack of expression and evoked emotions of disgust. Participants stated that the lack of expression came from ``A lifeless representation,'' ``Gray colors without expression,'' and ``It is only an object, so I didn't like it, has no emotion.''

\section{Conclusion}
\label{sec:conclusion}
AI is changing our everyday lives and also art creation. There are controversial discussions in the art community about the new definition of art, authorship, and creativity following increasingly widespread experimentation with AI from using off-the-shelf algorithms to generate final artworks to human intervention at every step of the AI process. AI art is firmly establishing its place in the art world through museum exhibitions and auction sales.
\par Motivated by this background, in this study, we aim to understand generative AI models as a new artistic medium by using deep neural net representations for style. More specifically, we aim to reveal the constellations of AI-generated art within the context of human art heritage in the computational space and visualize, observe, and statistically analyze them. The constellations are interpreted from various semantic aspects encompassing the fundamental art principles in art history and emotional experience finally delineating the right current state of machine art.  
\par In the analysis, we found that AI-generated art basically simulates the most typical subjects in training data—in our case, landscape, portrait, and geometric abstracts often drawn from the 1800 - 2000 period samples in WikiArt—but the outcomes often consist of bizarre shapes, distorted human faces and bodies, and unrecognizable nature scenes especially when generated by GAN models. The incomplete figuration causes the bias toward the visual concepts related to modern period art in abstract and open forms, but also in OOD analysis, the deformed images were detected as significant outliers compared to human art images. The phenomenon might be caused by the GAN models' architectures and learning algorithms in practice, but the constraint comes to form a unique and distinctive visual style of AI-generative art. In the diffusion model DDPM, on the other hand, we observed its ability to produce machine art with realistic and delicate human and natural figures. DDPM samples are liked the most by the participants in our survey. 
\par Our results and empirical observations reveal current generative art's unique visual features. And, we understand that historically new art mediums have initiated new art movements or genres like the camera, which was introduced as a scientific instrument, gradually revealed its artistic potential. However, this study did not aim at positioning AI-generated art into a new abstract category. Rather than that, we demonstrated how AI-generated art is different from human art heritage based on various fundamental aspects of art history and aesthetics in fine neural net representations. 
\par  The AI-generated art samples shown in this study represent the case where art production is solely based on the machine's random generation without artists' curation.  However, the practical use of the ML generative model in the art world is beyond that; it would be the middle of somewhere between pure randomness and the full realization of the author's free will. In addition, in the developmental process of generative models, we have many choices for datasets and ML algorithms. Given all possibilities and upcoming new algorithms, ML-based generative models possess endless potential for visual art. This study was about the starting point of the infinite spectrum. The framework in this study will provide a baseline for art and ML communities to understand current and next-generation machine art. To facilitate future research, the codes for the models, analysis, and collected data will be released.

{\small
\bibliographystyle{ieee_fullname}
\bibliography{egbib}
}
\onecolumn
\newpage
\appendix
\label{appendix}

\section{Data Collection Interface} \label{data_c_supp}

\begin{figure}[t]
\centering
\includegraphics[width=0.65\linewidth]{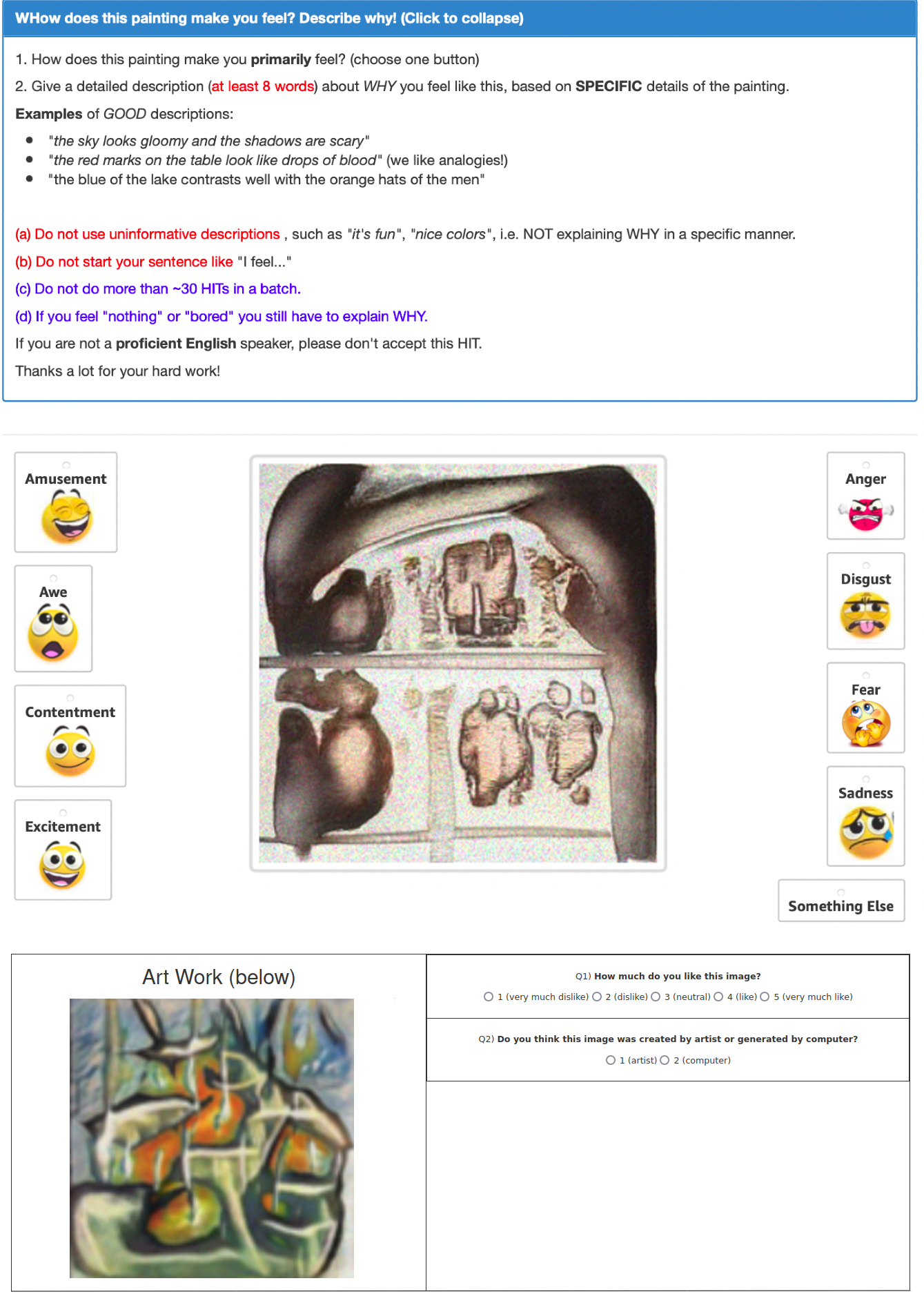}
\caption{User interfaces of the emotion experiment (top) and likability experiment(bottom). }
\label{fig:data_collection}
\end{figure}

\begin{figure*}[t]
\centering
\includegraphics[width=0.8\linewidth]{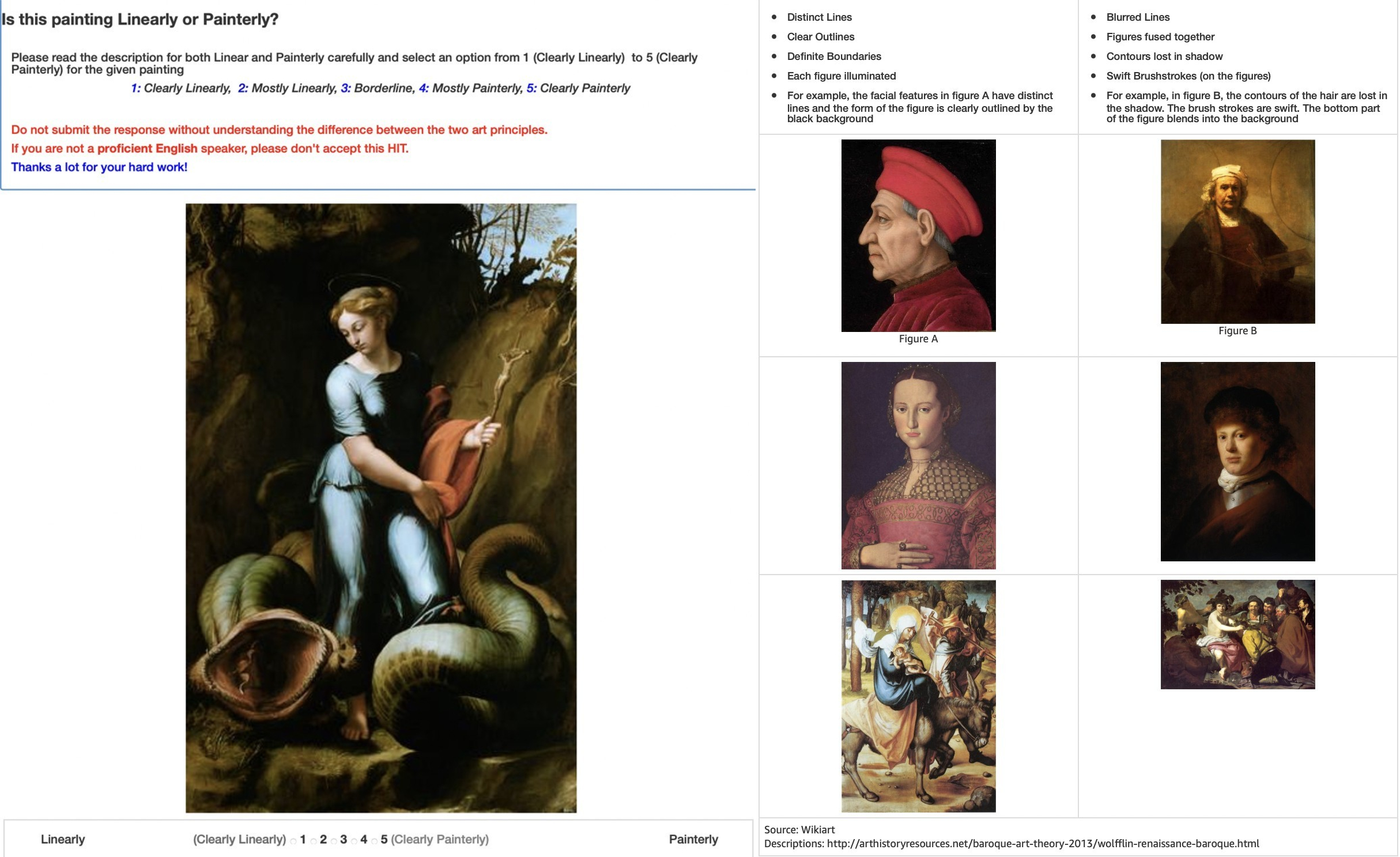}
\caption{Data Collection Interface of Wölfflin's principles presenting the form and examples used to train participants to classify the work as either linear or painterly.}
\label{fig:wofflin_exp}
\end{figure*}

We present the user interface utilized for collecting crucial data, including W\"olfflin's Principles, emotional responses, and Turing test data. Figure~\ref{fig:data_collection} and Figure~\ref{fig:wofflin_exp} visually illustrate the user interface for these tasks. 

The user interface design played a significant role in ensuring the effectiveness and accuracy of data collection. For the W\"olfflin's principles task (subsection 3.2.1 of the main paper), participants were prompted to identify the underlying principles depicted in the artworks after providing detailed instructions to understand the principle. The emotional responses task (subsection 3.2.5 of the main paper) involved presenting participants with various generated art pieces and human art, prompting them to select the emotions evoked by each piece. Lastly, the Turing test data collection included presenting participants with a mix of generated and human art pieces, challenging them to distinguish between the two.

We emphasize that the user interface was designed to be intuitive and user-friendly, minimizing any potential biases that could influence the data collection process. The results obtained from this comprehensive data collection methodology provided valuable insights into the effectiveness and capabilities of the generative models in evoking diverse emotions and artistic principles.

\section{Style Classification and Time Correlation} \label{class_supp}

\begin{table}[h!]
    \centering
	\scalebox{1}{
		\begin{tabular}{@{}lcc@{}}
			\toprule
			\multicolumn{1}{l}{Architecture} & trained from scratch &pre-trained \& fine-tuned \\
			\midrule
			\multicolumn{1}{l}{ResNet50}    & 27.80\% & \textbf{64.41\%} \\
			\multicolumn{1}{l}{ResNet50+2}  & 31.81\% & 60.00\% \\
			\multicolumn{1}{l}{ResNet101}   & 29.16\% & 64.13\%  \\
			\multicolumn{1}{l}{ResNet101+2} & 29.24\% & 64.27\%  \\
			\multicolumn{1}{l}{VGG16} & 31.48\% & 55.88\%  \\
			\multicolumn{1}{l}{VGG16+2} & 28.62\% & 56.91\%  \\
			\multicolumn{1}{l}{ViT-S} & \textbf{33.92\%} & 60.03\%  \\
			\multicolumn{1}{l}{ViT-B} & 32.68\% & 55.75\%  \\
			\multicolumn{1}{l}{ViT-L} & 33.81\% & 56.93\%  \\
   
			\bottomrule
		\end{tabular}
	}
\caption{The table displays each model's average accuracy per class. Pretrained and fine-tuned ResNet50 architecture shows the highest art-style classification accuracy.}
    \label{tb:classification}
\end{table}

\subsection{Style Classification}

Various convolutional and transformer-based networks are trained for the task of style classification. Following the training setting in \cite{elgammal2018shape}, we compared the two cases of (1) trained from scratch and (2) pretrained+fine-tuned from ImageNet models. For CNNs, VGG-Net16, ResNet50, ResNet101, and the other extended three ConvNet models are tested after adding the two hidden layers of 1024 and 512-dimensional nodes on the top of the original models. For transformer models, ViT-S, B, and L are considered.

These networks are trained on WikiArt. The dataset consists of more than 80,000 paintings belonging to 27-period styles ranging from the eleventh to the twenty-first century. In our experiments, the number of classes is reduced to 21 art classes by merging similar style classes; new-realism, contemporary-realism, and realism were merged, analytical-cubism, cubism, and synthetic-cubism were merged, and action-painting was merged with abstract-expressionism. The data is split into training (80\%), validation (10\%), and testing sets (10\%). Table \ref{tb:classification} presents the average classification accuracy per class of the models. As expected, the pre-trained and fine-tuned networks achieved significantly better results than their counterparts trained from scratch for all networks. Pretrained and fine-tuned ResNet50 architectures achieved the highest art-style classification accuracy.

\begin{wraptable}[21]{r}{0.45\textwidth}
\vspace{-2\baselineskip}

\centering
    \scalebox{1}{
    \begin{tabular}{@{}lc@{}}
    \toprule
        Architecture & \textbf{PCC}\\ 
        \midrule
        ResNet50 & \textbf{0.625}\\ 
        ResNet50+2 & 0.366 \\ 
        ResNet101 & 0.537\\ 
        ResNet101+2 & 0.503\\ 
        VGG16 & 0.462\\ 
        VGG16+2 & 0.422\\ 
        ViT-S & 0.542 \\
        ViT-B & 0.536 \\
        ViT-L & 0.489 \\
        SG1 Disc. & 0.184 \\ 
        SC1 Disc. & 0.316 \\ 
        SG2 Disc. & 0.224 \\ 
        SC2 Disc. & 0.320 \\ 
        CW1 Disc. & 0.385 \\ 
        CW2 Disc. & 0.286 \\ 
        \bottomrule
    \end{tabular}}
        \caption{
The maximum absolute Pearson's Correlation Coefficient (PCC) of the first 30 PCA components: correlation between human art's extracted features and year of making. The maximum PCC is ResNet50 followed by
ViT-S and ResNet101.} 
    \label{tb:time}
\end{wraptable}

\subsection{Time Correlation}
The high correlation of spatial arrangements in neural nets with time is the prior knowledge \cite{elgammal2018shape,kim2018artprinciple} we ground in our time analysis. As a preliminary step, we computed the PCC (Pearson's Correlation Coefficient) of principal components of our neural net features with the year of the artwork. In Table~\ref{tb:time}, adding the two layers on all CNN models caused a decline in PCC. The reduced dimension limits the representation power for the subtle visual information in artworks, so hard to capture the smooth visual transition over time. We also noted increased correlation with discriminators of both Style\textbf{GAN}-1 and Style\textbf{GAN}-2 architectures with the integration of the losses from CAN and CWAN. While the discriminators in Style\textbf{GAN}s are trained only to follow the original distribution of human art, the creative models—CAN and CWAN—are trained to be similar to human art but also not to be close to any of human art styles. Hence, the creative loss needs to know and leverage the information of period style to find a new visual space unoccupied. This can explain the high-time correlation results in the discriminators with CAN and CWAN in Table \ref{tb:time}. The maximum PCC is ResNet50 followed by ViT-S and ResNet101.

\subsection{Time Analysis from Different Networks}\label{timecorrelation}
\begin{figure*}[t]
\begin{subfigure}{.47\textwidth}
  \centering
  \includegraphics[width=\linewidth]{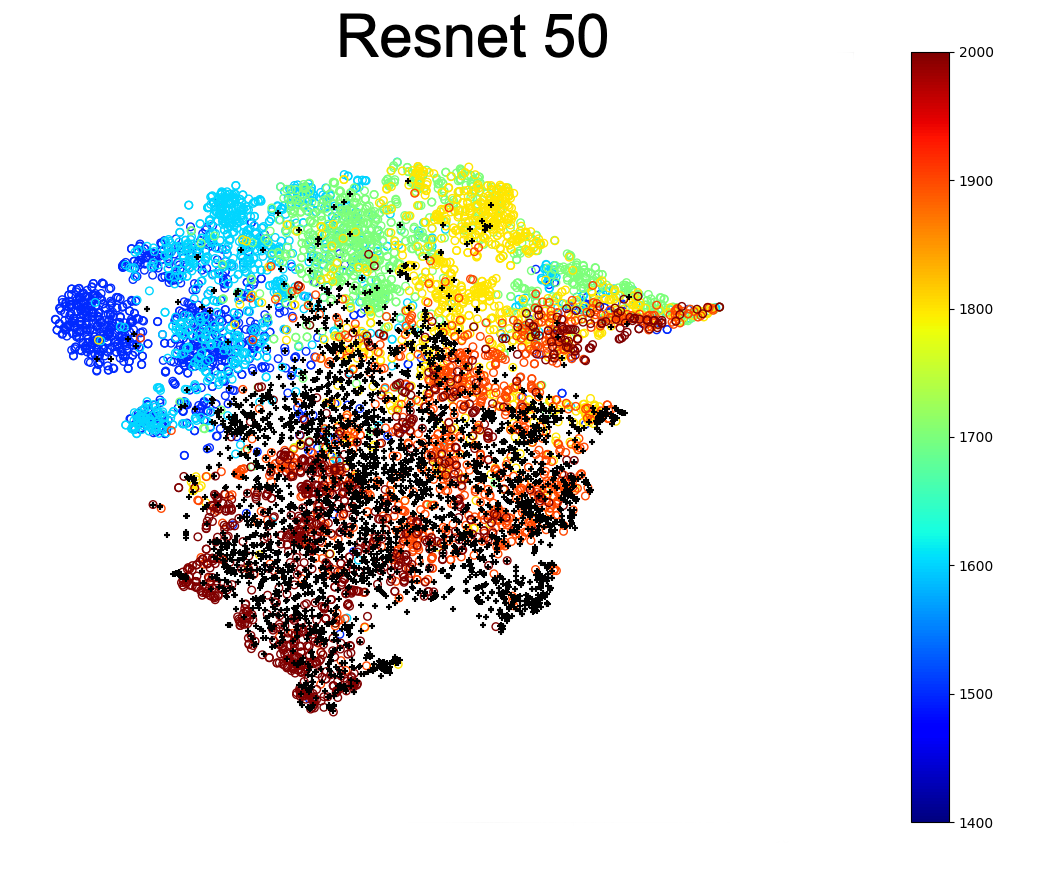}
\end{subfigure}%
\begin{subfigure}{.47\textwidth}
  \centering
  \includegraphics[width=\linewidth]{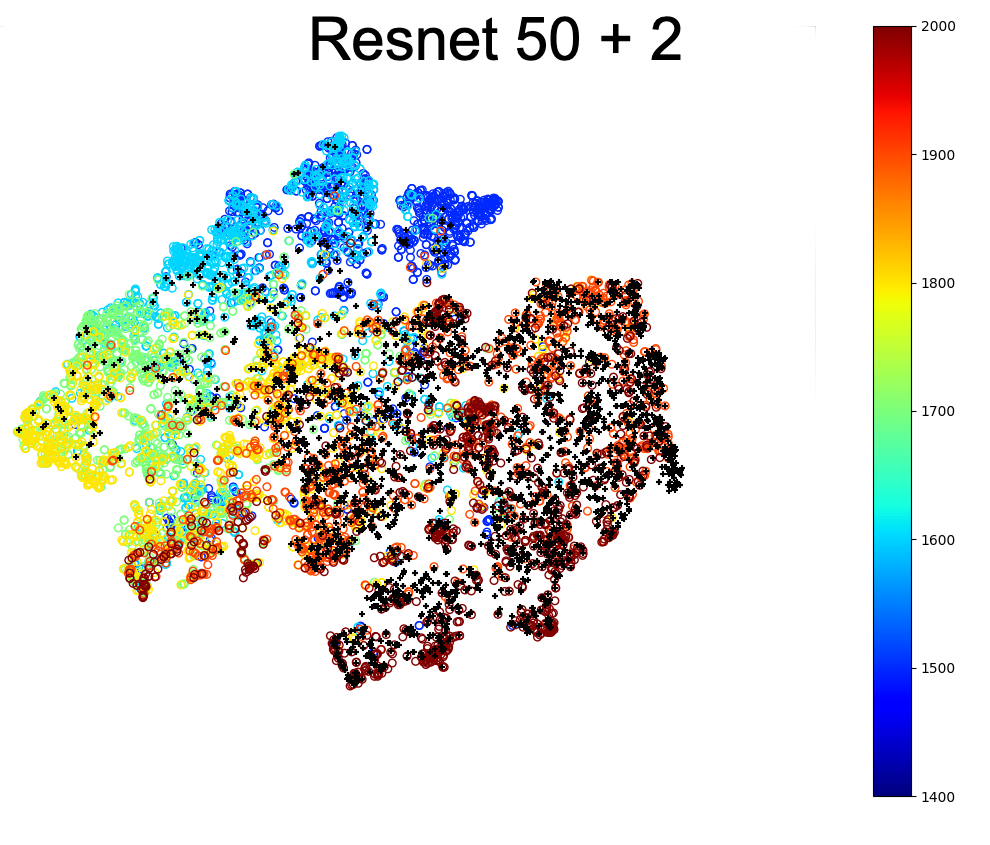}
\end{subfigure}
\begin{subfigure}{.47\textwidth}
  \centering
  \includegraphics[width=\linewidth]{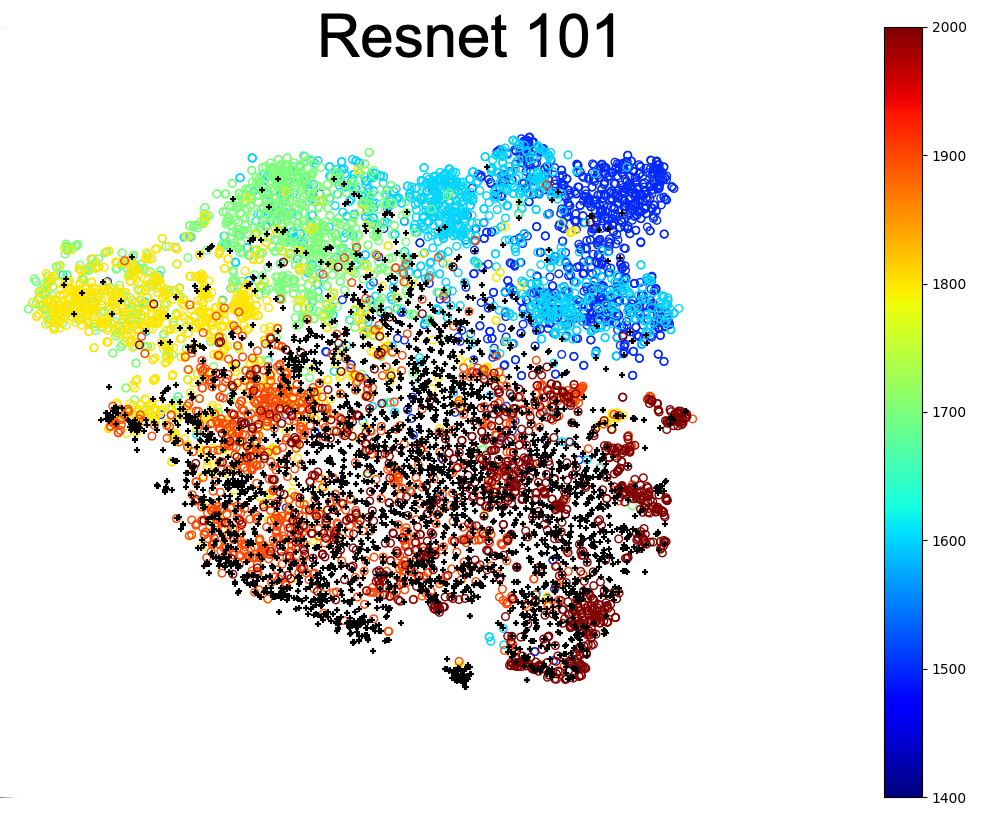}
\end{subfigure}
\begin{subfigure}{.47\textwidth}
  \centering
  \includegraphics[width=\linewidth]{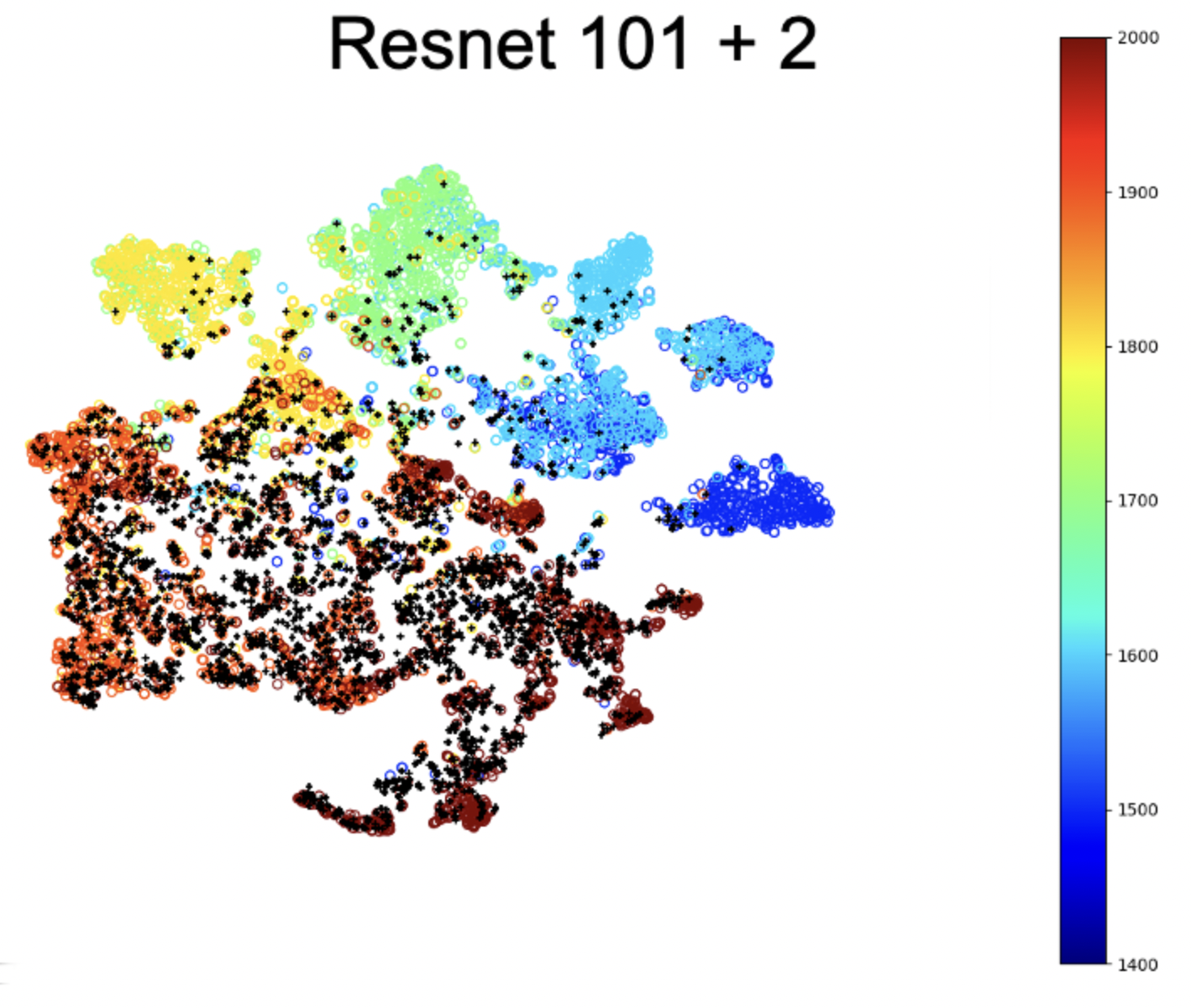}
\end{subfigure}
\caption{For ResNet50, ResNet50+2, ResNet101, and ResNet101+2, the t-SNE visualizations of features of generated art are overlayed over human art. The black markers are points corresponding to generated art.}
\label{fig:res_tsne_supp}
\end{figure*}

As done in Section 4.4 of the main paper, we overlay the t-SNE visualization from generated images onto human images. The features used in the main paper are from ResNet50~\cite{resnet}, we show results on feature representations from ResNet50, ResNet101, and their corresponding +2 models(ResNet50+2 and ResNet101+2) to compare how well the +2 models separate the human art in the time dimension and how the generated art is overlayed on the human art features. For ResNet50 and 101, the high correlation with time tells that these models can separate the human artworks based on their time period, and from Figure~\ref{fig:res_tsne_supp}, it can be seen that both these models allot generated art in the region close to modern art. The same trend is observed for the +2 models which show a slightly less correlation with the time dimension for human art. 

\clearpage

\section{W\"olfflin's Principles} \label{expaned_woff_supp}
\begin{figure}[t]
    \centering
    \includegraphics[width=1.0\linewidth]{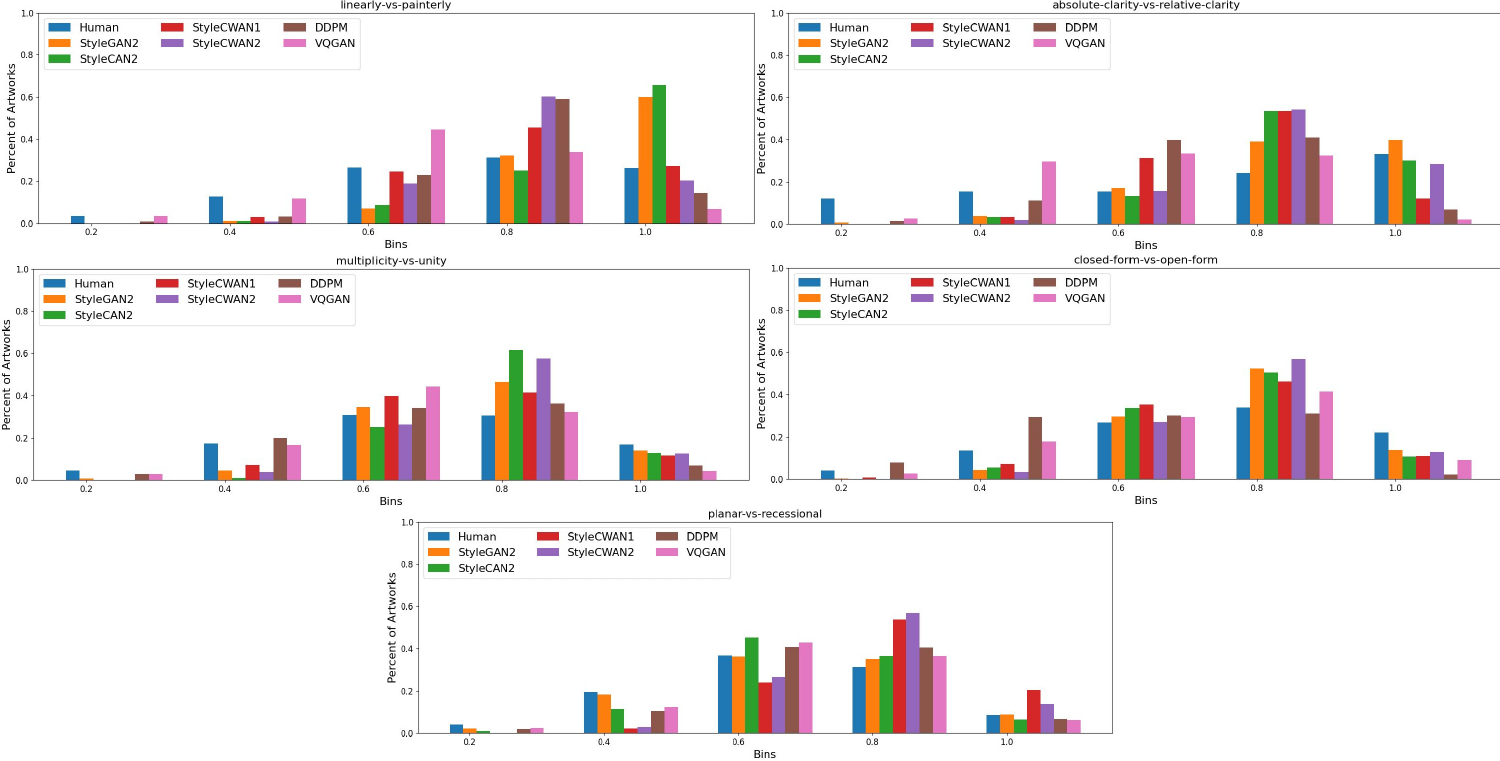}
    \caption{The distribution of all five Wölfflin's principles values: the bins represent the values from the left value to the right value. For example, the 0.2 bin represents the percentage of artworks that have values between 0 and 0.2. In this figure, AI-generated art is less uniform across the principle concepts than human art. Machine artworks are highly populated on the right side concepts more in general.}
    \label{fig:ratio_wofflin_supp}
\end{figure}

\begin{figure*}[t]
    \centering
    \includegraphics[width=\linewidth]{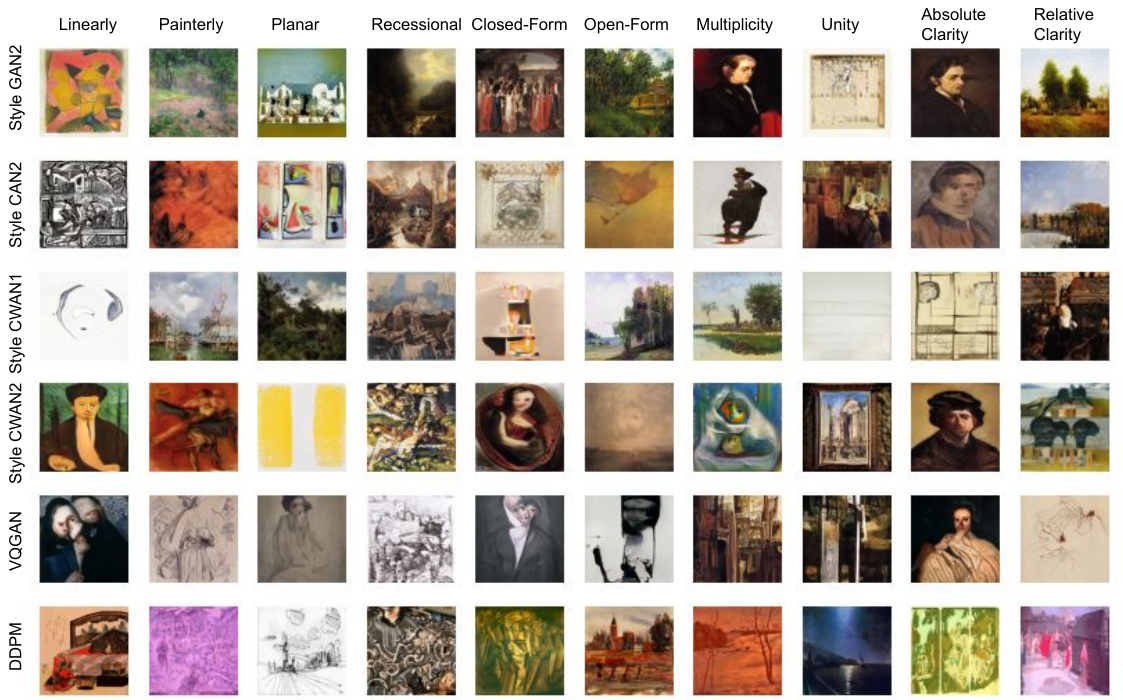}
    \caption{Generated art using StyleGAN2, StyleCAN2, StyleCWAN1, StyleCWAN2, VQGAN, and DDPM architectures for the different Wölfflin's principles.}
    \label{fig:avg_woff}
\end{figure*}

In Figure~\ref{fig:ratio_wofflin_supp}, we provide a detailed version of Figure 2 from the main paper for the analysis done in sub-section 4.1 of the main paper. We also present illustrative examples in Figure~\ref{fig:avg_woff} showcasing the highest values obtained for specific W\"olfflin's principles across different generative models. These examples serve as visual representations of generative art's ability to embody and emphasize particular artistic principles.

Each example in Figure~\ref{fig:avg_woff} highlights a distinct W\"olfflin's principle, effectively demonstrating how various generative models excel in incorporating and expressing different artistic attributes. By presenting these examples, we aim to provide a clear visual understanding of how each generative model can capture and manifest specific artistic characteristics.

\newpage

\clearpage

\section{General Art Principles} \label{proxy-analysis_supp}

\subsection{Experimental Procedures}
This section explains in detail normalization and hypothesis testing in statistical analysis on proxy space for comparison of population means between human and generated art. Lastly, we will introduce OODness-proxy (\textbf{O}ut-\textbf{O}f-\textbf{D}istribution) to measure the significance of a generated sample value compared to human art.

\subsubsection{Normalization by Human Art Statistics}
For proxy analysis, two proxy embeddings $R\in \mathbb{R}^{m \times 15}$ and $G \in \mathbb{R}^{n \times 15}$ are collected respectively as human and generated samples; the numbers of samples are $m$ and $n$.  To have unified scales for comparison, the two embeddings are normalized by the mean and standard deviation of $R$: $\mu_{R} \in \mathbb{R}^ {15}$ and  $\sigma_{R} \in \mathbb{R}^{15}$. All analyses on proxy space started from the normalized embeddings: $R'= \frac{R-\mu_{R}}{\sigma_{R}}$ and $G'= \frac{G-\mu_{R}}{\sigma_{R}}$, assuming the repetitive expansions of $\mu_{R}$ and $\sigma_{R}$ over the $m$ and $n$ samples. By normalization, the 15-means over the $m$ samples of $R'$ got exactly aligned with zero for all the 15 visual concepts, while the generated centers varied around zero depending on the relation of generated art to human art for different visual concepts. 

\subsubsection{Hypothesis Testing}
To compare the centers of human and generated arts after normalization, a hypothetical test is set as below.  
\begin{eqnarray*}
    H_{0}: \mu_{R} = \mu_{G}\\ 
    H_{a}: \mu_{R} \neq \mu_{G} \\ 
\end{eqnarray*}
The null hypothesis is rejected when the test statistic: $|Z|>3.0$ and $p$-value: $0.0026$. The test statistic is  $Z=\frac{\bar{X_{G}} -\bar{X_{R}}}{\frac{S_{G}}{\sqrt{n}}}$, where  $\bar{X_{G}}$ and $\bar{X_{R}}$ are the sample means for generated and human artworks, $S_{g}$ is the sample standard deviation of generated samples, and $n$ is the number of the samples. ~\cite{mendenhall2016statistics} are referenced to set up our experiments.

\subsubsection{OODness-proxy}
The smaller tail probability of a generated sample on the normalized space by human statistics implies a significant visual deviation from general human art. Let $p$ be the tail probability, and $x$ be an arbitrary instance value in $G'$ for $i$-th sample and $j$-th concept, and $n(x)$ is a normal distribution with $\mu=0$ and $\sigma=1$: $n(x) = \frac{1}{\sqrt{2\pi}}\exp^{-x^{2}}$ Then,OODness-proxy is computed as below. It measures how a generated sample is outlied visually from the center of human art on proxy space.  
\begin{align*}
p = min (\int_{-\infty}^x \! n(x) \, \mathrm{d}x,\int_{x}^{-\infty} \! n(x) \, \mathrm{d}x)\\
\textrm{OODness} = -\log{p}
\label{OODness}
\end{align*}

\newpage

\subsection{Hypothesis Testing Results for Individual Generative Models} \label{hypo_test_supp}
\begin{table*}[h]
\centering
\begin{adjustbox}{width=\textwidth}
\begin{tabular}{c| ccccccccccccccc}
\toprule
& \multicolumn{15}{c}{\textbf{$Z$- statistics for 15 Visual Concepts}} \\ \midrule
\textbf{models} & non-representational & representational & geometric & abstract & planar 
& closed & open & rough  & perspective & broken 
& thin & flat & distorted & linear & all \\ \midrule
        
Style\textbf{GAN}-1 & \textcolor{blue}{5.78} & \textcolor{blue}{-5.12} &\textcolor{blue}{4.34} & \textcolor{blue}{6.92} & \textcolor{blue}{5.65} & 
        \textcolor{blue}{-4.17} & -0.3 & \textcolor{blue}{6.43} & \textcolor{blue}{-6.95} & 
        \textcolor{blue}{6.27} & \textcolor{blue}{-6.12} & \textcolor{blue}{6.84} & \textcolor{blue}{6.25} & \textcolor{blue}{3.73} & \textcolor{blue}{6.45}\\ 
        
        Style\textbf{GAN}-2 & -0.5 & 0.79 & -1.39& -1.11 & -0.37 &
        -1.25 & 1.3 &0.7 & 0.35 &-0.57 &
        1.59 & -0.15& -0.22 &-0.74 & -0.96\\ 
        
        StyleCAN1  & 2.53 & -2.16 & 2.15& \textcolor{blue}{3.6} & \textcolor{blue}{3.44} &
    \textcolor{blue}{-3.14} & -0.59 & \textcolor{blue}{4.85} & \textcolor{blue}{-4.29} & \textcolor{blue}{4.14} & 
       -1.55 & \textcolor{blue}{4.86} & \textcolor{blue}{4.92} & \textcolor{blue}{3.81}& \textcolor{blue}{3.04} \\ 
        
        StyleCAN2 & \textcolor{blue}{4.38} & \textcolor{blue}{-4.24} & \textcolor{blue}{3.29} & \textcolor{blue}{3.93} & \textcolor{blue}{3.79} &
        -2.39 & 1.17 & \textcolor{blue}{3.01} & \textcolor{blue}{-3.66} & 2.04 &
        -2.77 & \textcolor{blue}{3.4} & 2.69 & 1.47 & \textcolor{blue}{4.48}\\ 
        
       StyleCWAN1& 1.38 & -1.27 & 1.18 & 2.63 &\textcolor{blue}{3.26} & \textcolor{blue}{-3.01} & -1.68 & \textcolor{blue}{4.72} & -2.96 &    
        \textcolor{blue}{3.72} & -0.91 & \textcolor{blue}{3.92} & \textcolor{blue}{3.97} & 2.62 & 2.14 \\ 
        
        StyleCWAN2 & 2.26 & -1.97 & 1.13& 3.05 & 2.3 &
        \textcolor{blue}{-6.04}& \textcolor{blue}{3.11} & \textcolor{blue}{5.66} & \textcolor{blue}{-3.94} & \textcolor{blue}{4.03} &
        -2.05 & \textcolor{blue}{3.7} & 2.87 & 1.3 & 2.74 \\ 
               
        VQ-GAN & 0.08 & 1.07 & -2.75 & 0.07 & -0.47 &
       -1.59 & 1.48 & 2.24 & -0.8 & -0.61 & 
        \textcolor{blue}{-3.15} & 0.15 & 1.15 & \textcolor{blue}{-3.66} & 0.17\\ 
        
       DDPM & \textcolor{blue}{19.0} & \textcolor{blue}{-18.29} & \textcolor{blue}{17.71} & \textcolor{blue}{18.89} & \textcolor{blue}{17.5} & \textcolor{blue}{-11.03} & \textcolor{blue}{8.17} & \textcolor{blue}{14.76} & \textcolor{blue}{-18.8} & \textcolor{blue}{11.51} & \textcolor{blue}{-9.41} &
        \textcolor{blue}{17.78} & \textcolor{blue}{5.89} &  \textcolor{blue}{8.69} & \textcolor{blue}{19.63}\\ 
 
\bottomrule
\end{tabular}
\end{adjustbox}
\caption{For all generative models, we computed  $Z$-statistics. Even though different models had different significant visual concepts, the signs of the significant statistics were consistent across the different models except for ``linear" on VQ-GAN, and were matched to the direction of the visual changes from classical to modern art. For example, modern paintings are more ``non-representational" and ``geometric", so positive statistics while less ``representational" and  ``perspective" than classical human art, so got negative statistics on the concepts. }
\label{tb:zstatistics}
\end{table*}

\section{Qualitative Examples in Turing Test} \label{turingtest}
\begin{figure*}[h]
\centering
\includegraphics[width=\linewidth]{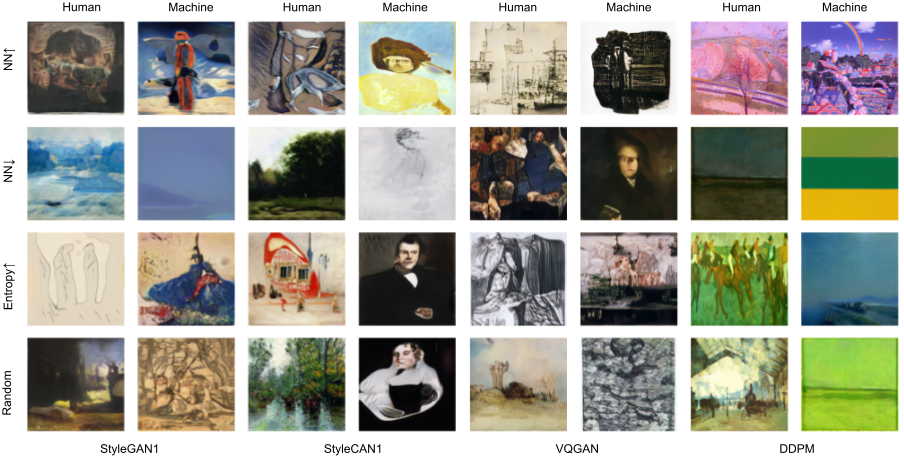}
\caption{AI-generated art using Style\textbf{GAN}-1, Style\textbf{CAN1}, VQ-GAN, and DDPM architectures (from left to right respectively), which is most likely believed by users to be human-made and which are believed to be generated by a machine. }
\label{fig:turing}
\end{figure*}

We show some examples in Figure~\ref{fig:turing} from AI-generated art that is selected by users as most likely to be drawn by humans and some that are most likely AI-generated.

\end{document}